\documentclass[lettersize,journal]{IEEEtran}
\usepackage{amsmath,amsfonts}
\usepackage{algorithmic}
\usepackage{array}
\usepackage{textcomp}
\usepackage{stfloats}
\usepackage{url}
\usepackage{verbatim}
\usepackage{graphicx}
\usepackage{lipsum}
\usepackage{subcaption}
\def\BibTeX{{\rm B\kern-.05em{\sc i\kern-.025em b}\kern-.08em
    T\kern-.1667em\lower.7ex\hbox{E}\kern-.125emX}}
\usepackage{balance}
\usepackage{booktabs}

\begin{document}

\title{A Multimodal Tiltwing Framework for Bioinspired Aerial Robots}

\author{Krispin C.V. Broers and Sophie F. Armanini
\thanks{Krispin C.V. Broers (corresponding author), k.broers24@imperial.ac.uk, Department of Aeronautics, Imperial College London, London, SW7~2AZ, UK.}
\thanks{Sophie F. Armanini, Department of Aeronautics, Imperial College London, London, SW7~2AZ, UK.}}

\markboth{Preprint June 2026}{}

\maketitle

\begin{abstract}
Tailless flapping-wing micro-aerial vehicles (FWMAVs) mimic the impressive flight performance of hummingbirds, utilising unsteady aerodynamic effects. However, existing designs are still limited and purpose-built with a restricted flight envelope and poor endurance.
We therefore propose an adaptable tiltwing framework enabling bioinspired aerial robots to switch between hovering flight, high-speed directional flight, and energy-efficient gliding flight.
The proposed framework utilises thrust vectoring with a wide actuation range via two fully independent propulsion units, each flapping a single wing, for effective control and enhanced manoeuvrability. For this, we developed a hybrid Scotch-yoke-based flapping mechanism that ensures a symmetric motion profile with a modular design guaranteeing an arbitrarily wide flapping angle to exploit the lift-enhancing clap-and-fling effect.
Additionally, we implemented a passive wing-rotation mechanism, which, in combination with our dual-wing thrust-vectoring approach, allows unprecedented wing-design freedom, unlocking potential for precise optimisation.
A contactless leading-edge tracking sensor provides accurate feedback on the wing's orientation and, in the gliding mode, enables dihedral-angle control, augmenting the active wing-pitch control.
Extensive testing of a propulsion unit was conducted with a six-axis force/torque sensor, demonstrating the flapping mechanism's performance while optimising transmission efficiency and the passive wing-pitch mechanism. At full throttle, the average lift force generated by a single wing, flapping with a 188º amplitude, was 21.1\,gf for a small 3.1\,g 1S BLDC motor. Additional tests covering the full range of the wide-angle tilting capability showed an effective thrust-vectoring control architecture with a linear and symmetric response curve of the moments generated.
\end{abstract}


\section{Introduction}
Birds and insects demonstrate exceptional flight performance across a wide flight envelope, with examples including hovering hummingbirds, agile swifts, gliding pigeons, and soaring seagulls. In recent decades, with advances in microelectronics and aerial robotics, interest in replicating the manoeuvrability and efficiency of low-Reynolds-number flight found in nature has increased rapidly. The category of small unmanned flying robots was defined as micro aerial vehicles (MAVs) by DARPA in 1997 \cite{Gundlach2012}, with the first designs of fixed-wing MAVs emerging not long after, e.g. the AeroVironment Black Widow by Grasmeyer and Keennon~\cite{Grasmeyer2001}. 
A more direct bioinspired approach led to the creation of flapping-wing MAVs (FWMAVs), which utilise oscillating wings for lift and thrust production. These can be categorised into tailed FWMAVs (e.g.~\cite{Pornsin-Sirirak2001,DeCroon2012}), inspired by large birds, or tailless FWMAVs, inspired by insects and hummingbirds.
Tailed designs, however, can lack control authority at lower speeds, struggle in hover, and are more susceptible to perturbations \cite{RafeeNekoo2025}.
To mitigate these disadvantages and provide exceptional manoeuvrability, as displayed by hummingbirds, tailless configurations were proposed. One of the most prominent and capable tailless FWMAVs was the 19\,g Nano Hummingbird by Keennon et al.~\cite{Keennon2012}. Within the tailless designs, different configurations exist: Two-winged designs with a single motor are most prevalent, such as the KU-Beetle by Phan et al.~\cite{Phan2017}, the Colibri by Preumont et al.~\cite{Preumont2021}, the Texas A\&M hummingbird by Coleman et al.~\cite{Coleman2017}, and the BionicBee from Festo~\cite{Festo2024}.
These two-winged designs more closely follow examples found in nature and feature less complex actuation systems due to the often mechanically linked wing pair \cite{Phan2019}.
However, they require high flapping frequencies and large stroke amplitudes to produce sufficient lift \cite{Phan2019}.
Another common design features two motors and an X-wing configuration; they provide enhanced stability due to two individually actuated wing pairs providing augmented lift production via the clap-and-fling effect \cite{Nguyen2018}. 
This unsteady lift-enhancing mechanism is found in various flying insects, notably in butterflies, where at the end of a stroke the wings touch, i.e. the "clap", pushing out the fluid between them, followed by a release, i.e. the "fling", that increases circulation by sucking in surrounding fluid~\cite{Weis-Fogh1973, Brodsky1991, Sane2003}.
The most notable example for the X-wing configuration was the tailless DelFly Nimble by Karasek et al.~\cite{Karasek2018}, which exhibited exceptional manoeuvrability. Other examples of two-pair X-wing designs are the NUS-RoboticBird by Nguyen and Chan~\cite{Nguyen2018}, the JT-Fly by Wu et al.~\cite{Wu2024}, and the X-fly by Wu et al.~\cite{Wu2026}.
These three designs proved especially capable in their control moment generation through the implementation of thrust vectoring, where the left and right wing pair are each independently actuated by a motor and a tilting servo.
As a configuration, the X-Wing design adds complexity due to the wing-pair actuation, which also restricts the wing design space, with the two wings in a pair needing to be joined.

Overall, the tailless avian-scale FWMAVs show tremendous potential, due to their agility, performance at low flight speeds, and safe propulsion system when compared to propellers or rotors. This would make an application in cluttered environments or around humans, animals and plants ideal. However, most designs are purpose-built solutions optimised to study a single flight mode or application. This is coupled with a lack of endurance, with many designs achieving less than 5~minutes of hover flight time (e.g.~\cite{Coleman2017,Nguyen2018,Karasek2018,Preumont2021}.
A recent study by Wu et al.~\cite{Wu2026} showed improvements, achieving 33.2~minutes of flight with an especially large 21.6\,g battery with a capacity of 1100mAh. Another approach consists of designing robots towards multimodality and energy conservation, such as temporarily entering an efficient perching state, as demonstrated with a gripper system on FWMAVs by Broers and Armanini~\cite{Broers2025}, and Zufferey et al.~\cite{Zufferey2022}. Another multimodal FWMAV that can enter a crawling mode as an additional type of locomotion on the ground was developed by Wu et al.~\cite{Wu2026}.

Fully exploiting the extensive flight envelope of FWMAVs would be a game-changer, substantially increasing the range of applications and conserving energy during flight through flexible adaptation of the robot's flight modes to its mission requirements.
To achieve this, we aim to merge the configurational simplicity and wing-design freedom of a tailless two-winged solution with an effective thrust-vectoring control architecture, only recently implemented on X-Wing configurations \cite{Nguyen2018, Wu2026}.
Our framework additionally proposes a wider actuation range for the thrust vectoring to provide the robot with aggressive manoeuvrability and the ability to adapt to different flight modes efficiently. This will also enable the decoupling of wing actuation and body attitude, allowing further control strategies leading to potentially new applications.
Besides hovering flight and omnidirectional high-speed flight, this two-winged design will also be able to perform an energy-conserving glide mode, which makes use of the wide range in wing articulation for control.
Furthermore, the proposed thrust-vectoring control architecture, which does not rely on wing deformation utilised in other two-winged robots (e.g.~\cite{Keennon2012,Phan2017,Roshanbin2017,Preumont2021}), creates a large wing design space providing opportunities for further studies on the enhancement of thrust production over the full flapping cycle.
Contrary to the X-wing or coupled two-wing designs, the proposed framework requires a leading-edge trajectory tracking system to synchronise the individually actuated wings. Similar systems were implemented by Tu et al.~\cite{Tu2020} and Guo et al.~\cite{Guo2024}, which use motor encoders on independent direct-driven dual-wing designs. Direct driving the wings, i.e. without a flapping mechanism, however, suffers from efficiency losses due to constant motor reversal~\cite{Guo2024}.
In our framework, this wing trajectory tracking system, in combination with the driving motor and flapping mechanism, will additionally serve as a closed-loop actuator for the wings' orientation during gliding flight.

Our main contributions to furthering the capabilities of aerial robots and expanding their flight envelope include the following:
\begin{itemize}
    \item Proposal of a novel design framework for a multimodal tiltwing aerial robot, capable of adapting to hovering flight, high-speed directional flight, and energy-efficient gliding.
    \item Demonstrating a novel approach to exploit effective thrust-vectoring control over a wide actuation range and decouple propulsion and body attitude through tilting wings.
    \item Development of a modular wide-angle flapping mechanism with a symmetric motion profile for two-winged designs, and an integrated wing-trajectory-tracking sensor, enabling the multimodality of the framework.
    \item Implementation of a passive system for wing rotation, which in combination with our thrust-vectoring actuation enables unprecedented wing-design freedom and unlocks the potential for precise optimisation.
    \item Conducted extensive testing of the framework, utilising a force/torque test rig to validate the viability of the transmission design, the control architecture over the full actuation range, and the multimodal approach.
\end{itemize}

In Section~\ref{designframework} we explain the architecture of the proposed multimodal tiltwing framework, followed by the design of the flapping mechanism, and our approach to wing position tracking.
This is followed by Section~\ref{multibody} where we present the multibody and aerodynamics model that helped the development of our framework and its components.
Section~\ref{testing} details the experiments conducted with a propulsion unit prototype quantifying general performance metrics and the control moments generated by our wide-angle thrust-vectoring design.
Finally, in Section~\ref{conclusion} a conclusion is given with an outlook at future work.

\section{Design Framework}\label{designframework}
At first, the framework was developed with its design goal of creating a multimodal aerial robot excelling at various flight modes, therefore extending the available flight envelope and enhancing mission capabilities. With this, possible control strategies for each flight mode of the proposed framework are presented. After defining the design requirements that enable the framework, a tailored flapping mechanism was developed
The requirements for this design were laid out, and a novel flapping mechanism tailored to this approach was developed. 
Lastly, a wing position sensor was developed for the transmission, enabling a key feature of the proposed framework with independent propulsion units for each wing and the ability to enter a glide mode.

\subsection{Multimodal Tiltwing Framework}
\begin{figure*}[tb]
    \centering
    \includegraphics[width=0.7\linewidth]{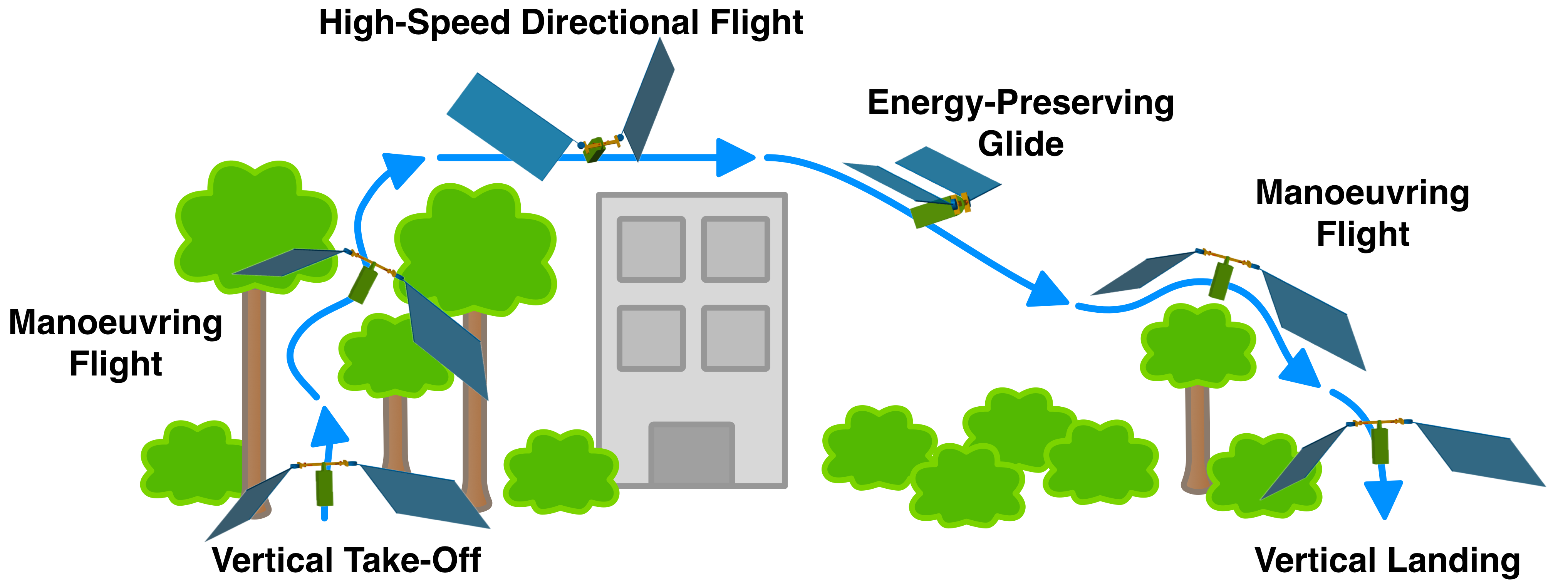}
    \caption{View of an exemplary mission utilising the adaptability of the framework.}
    \label{fig:mission}
\end{figure*}

The multimodal framework should be able to optimally adapt to the mission requirements, guaranteeing efficient flight. An exemplary mission that utilises the versatility is shown in Figure~\ref{fig:mission}.

This mission starts in a cluttered environment where exceptional manoeuvrability is required to avoid collisions, and a vertical take-off into hover is used. This is followed by manoeuvring flight in the hover mode until all obstacles are cleared. The robot then changes into the high-speed directional flight mode to effectively cover a longer distance. A unique feature of the proposed framework is the ability to intermittently switch into an energy-preserving glide mode, which can be used, for example, to efficiently approach the target location. To precisely land at this desired location, the robot once again switches into hovering mode and manoeuvres around obstacles, finishing the mission with a vertical landing sequence.

\begin{figure}[htb]
    \centering
    \includegraphics[width=0.9\linewidth]{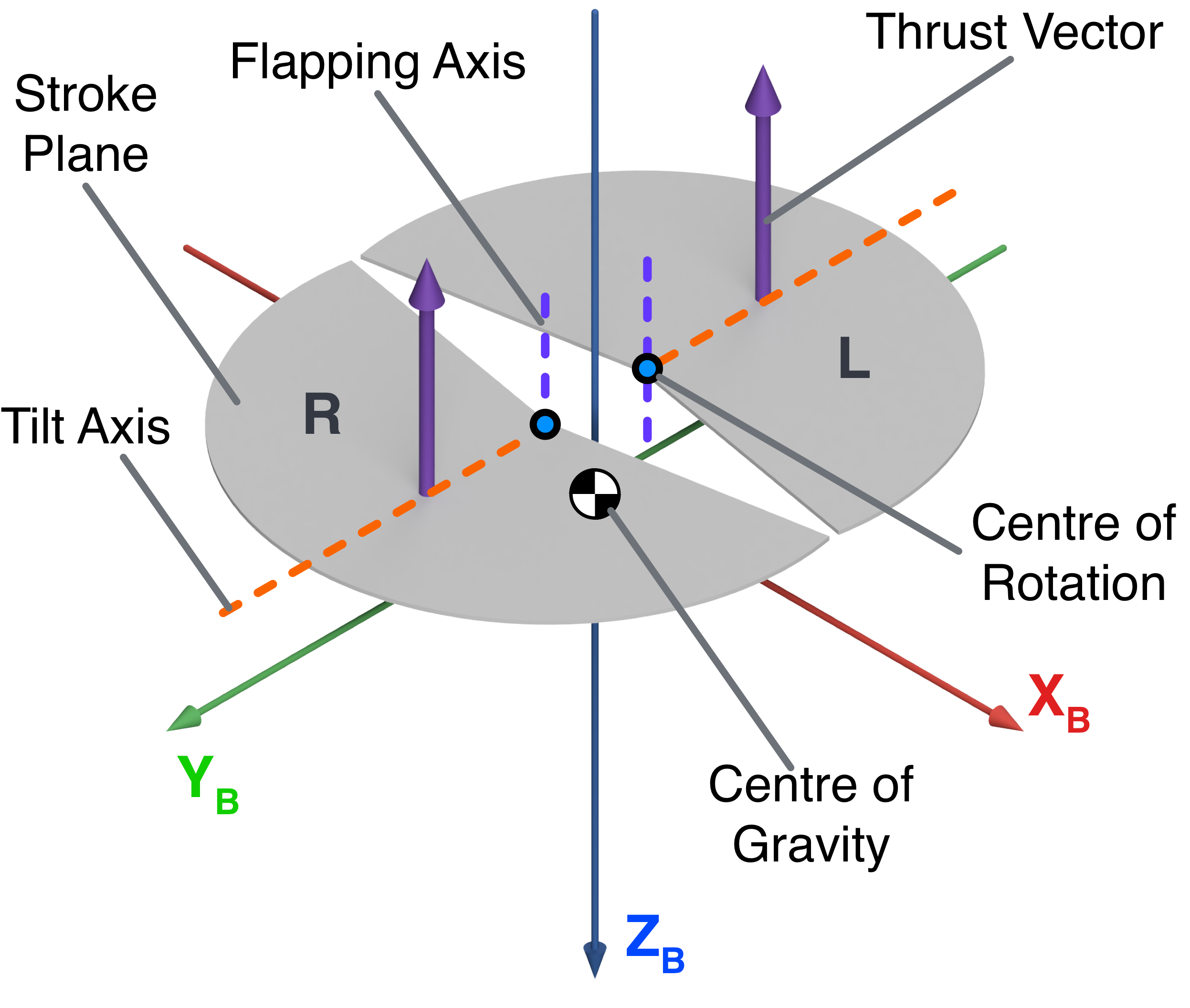}
    \caption{Functional view of the proposed framework, showing two independently-actuated tilting stroke planes in relation to the body-fixed frame. The stroke plane represents the area in which the leading edge of the wing travels around the flapping axis, generating a thrust vector that can be tilted around the tilt axis or scaled in magnitude with the flapping frequency for control.}
    \label{fig:fwcomponents}
\end{figure}

To achieve these kinds of missions and provide a new level of versatility through flight envelope exploitation, a two-winged tailless flapping-wing aerial robot design framework was developed. For this, the two wings are fully independent of each other and utilise each an efficient propulsion unit with a novel flapping mechanism optimised for two-winged operation, wing design freedom, and the exploitation of the lift-enhancing clap-and-fling effect.
To control the unstable tailless configuration and enable adaptability to multiple flight modes, such as hover, directional flight, and glide, a wide-ranging tilting mechanism for each of the two propulsion units is proposed. This enables the framework to utilise highly effective thrust vectoring for enhanced manoeuvrability and versatility in the control actuation for each mode. 
The tilting of the propulsion units around the spanwise axis of the wings (see Figure~\ref{fig:fwcomponents}), resulting in the tilting of the individual stroke planes, is envisioned to span a hemisphere, enabling the generation of large control moments and the adaptation to the gliding mode. 
In total, the framework consists of four fully independent actuators, i.e. two propulsion unit motors and two tilting servo-motors, providing granular control of the individual wings with a wide actuation range.

\begin{figure}[htb]
    \centering
    \includegraphics[width=\linewidth]{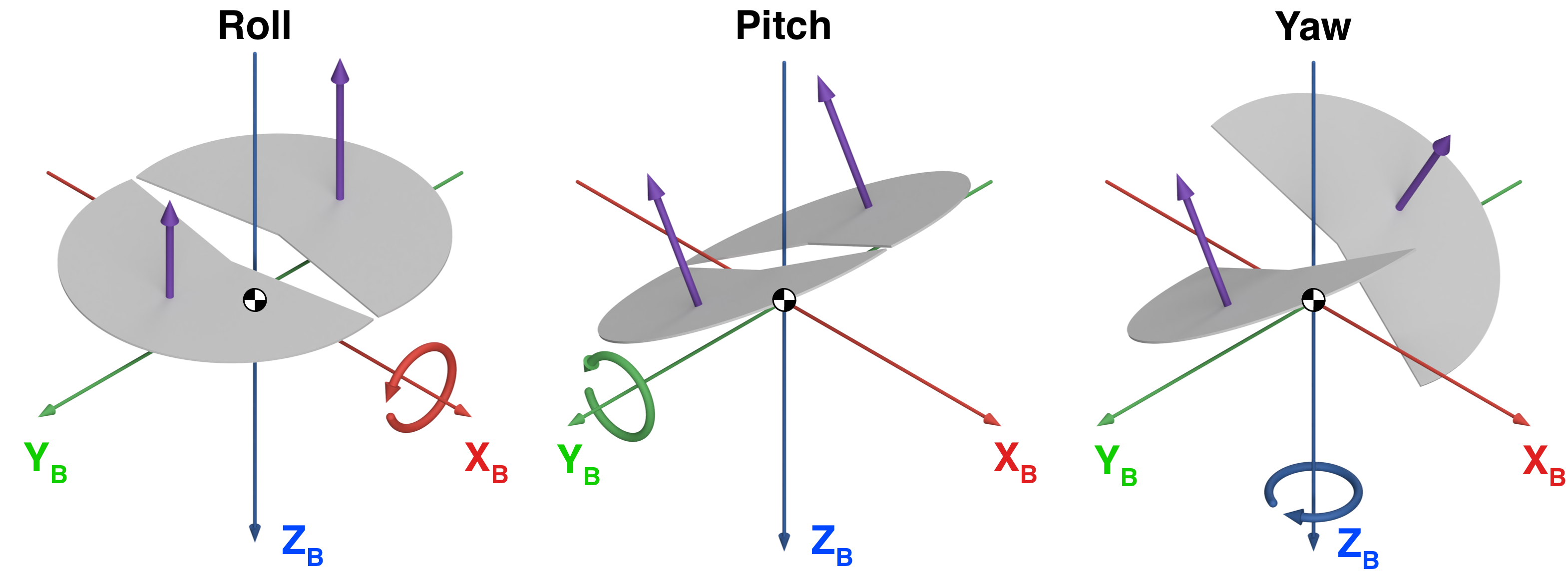}
    \caption{Control configuration for roll, pitch, and yaw in the hovering flight mode showing the thrust vectoring approach.}
    \label{fig:hoverctrl}
\end{figure}

For the \textbf{hovering flight mode}, the configuration is in a neutral position with both stroke planes of the wings in a horizontal orientation and the body axis $Z_B$ of the robot pointing to the ground. Control forces are generated in this mode based on thrust vectoring, via modulation of the flapping kinematics. With this, the main motors regulate the flapping frequency for the left and right propulsion unit and thus the magnitude of the cycle-averaged thrust vectors, while the tilting servo-motors control the stroke-plane tilt for the left and right wing and thus the orientation of each thrust vector.
This leads to the following proposed control architecture in hovering flight, as displayed in Figure~\ref{fig:hoverctrl}:
\begin{itemize}
    \item Roll: Differential motor speed between the left and right propulsion units, resulting in a differential in thrust vector magnitude.
    \item Pitch: Symmetric stroke-plane tilt of the left and right propulsion units, resulting in a collective thrust vector tilt.
    \item Yaw: Differential stroke-plane tilt between the left and right propulsion units, resulting in a differential thrust vector tilt.
\end{itemize}

\begin{figure}[htb]
    \centering
    \includegraphics[width=\linewidth]{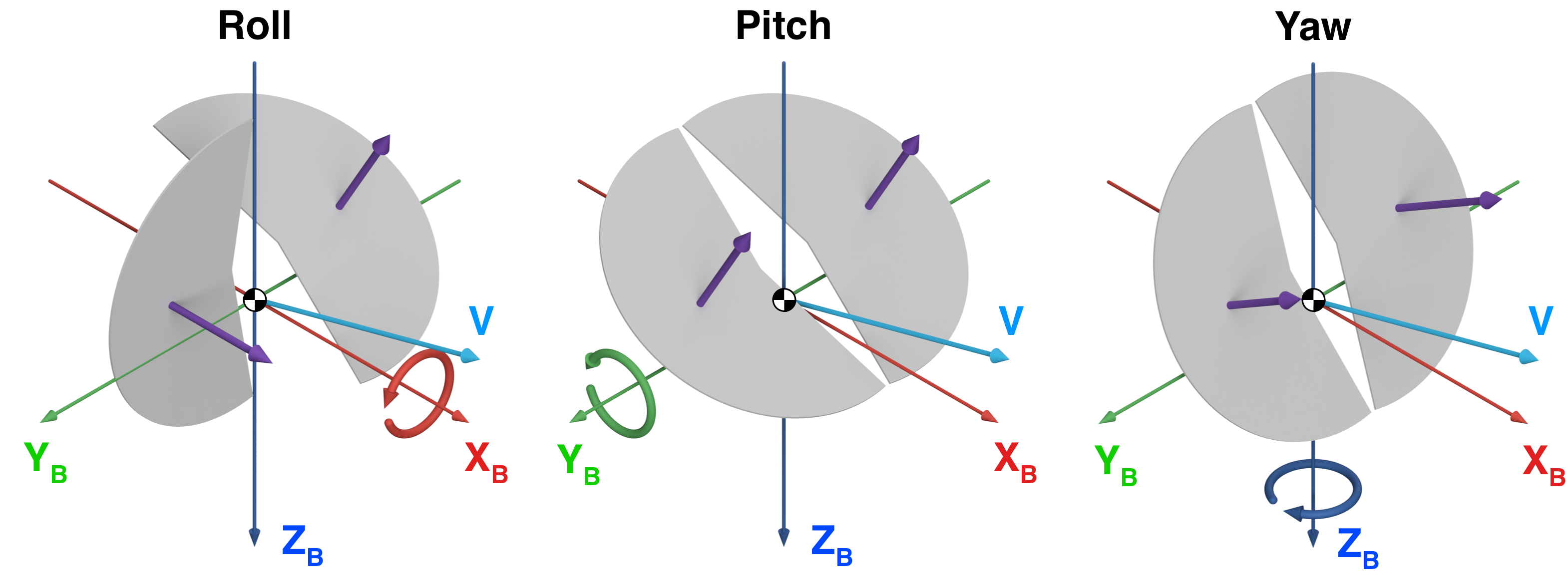}
    \caption{Potential control configuration utilising approach (B) for roll, pitch, and yaw featuring an upright body orientation in the high-speed directional flight mode, with the flight direction marker $V$.}
    \label{fig:fwdctrl}
\end{figure}

The versatility of the wide-angle thrust vectoring allows for different approaches to the \textbf{high-speed directional flight mode}, e.g. during forward flight.
A straightforward approach (A) would be to tilt the body orientation parallel to the flight direction and generate control moments around the body-fixed axes, similar to the configuration in the hover mode described above. In forward flight, this would result in the $-Z_B$-axis pointing forwards.
Another approach (B) would be to utilise the wide tilting angle to keep the robot's body always oriented such that the $Z_B$-axis points down, perpendicular to the ground. This could be used to maintain a given body attitude and improve body stabilisation and thus also stability of the attached payload. In this case, the stroke planes and thrust vectors of both wings would tilt forward, towards the flight direction, maintaining the heading with control enacted as follows (see Figure~\ref{fig:fwdctrl}):
\begin{itemize}
    \item Roll: Differential stroke-plane tilt between the left and right propulsion units, resulting in a differential thrust vector tilt.
    \item Pitch: Symmetric stroke-plane tilt of the left and right propulsion units, resulting in a collective thrust vector tilt.
    \item Yaw: Differential motor speed between the left and right propulsion units, resulting in a differential in thrust vector magnitude.
\end{itemize}

This approach, however, will introduce cross-coupling effects due to the location of the centre of gravity, which is positioned below the wing-attachment points and therefore the wing-tilt axis (see Figure~\ref{fig:fwcomponents}) for increased passive stability.
The optimal centre-of-gravity location and its effect on cross-coupling will need to be studied in future work. Additionally, a control-mixer dependent on heading and thus degree of stroke-plane tilt will need to be implemented to effectively enact control forces and minimise cross-coupling effects.

In general, our proposed framework allows the freedom to choose the best performing control approach to fit the specific needs of a desired mission profile.

\begin{figure}[htb]
    \centering
    \includegraphics[width=\linewidth]{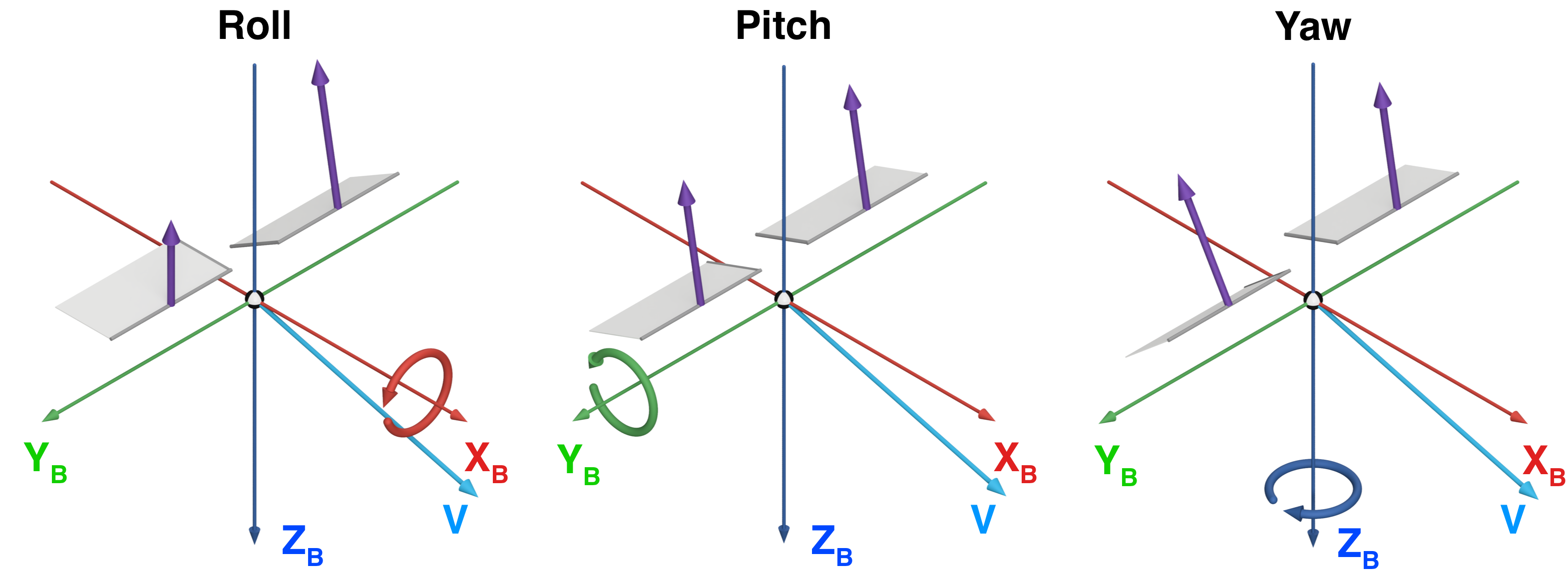}
    \caption{Potential control configuration for roll, pitch, and yaw in the gliding flight mode.}
    \label{fig:glidectrl}
\end{figure}

Lastly, for the energy-efficient and silent \textbf{gliding flight mode}, no thrust is produced, and the configuration acts similarly to a glider without a tail.
In this mode, active wing pitch control is utilised, which is achieved by combining the wide-angle tilting capability of the framework with the passive wing pitch mechanism. Lift and drag forces fixate each wing at the maximum position of its passive wing-rotation mechanism, with the wing tilt actively changing the incidence angle of each wing.
Additionally, in this mode, the motor-position control allows the flapping mechanism to change the wings' orientation, leading to a change in dihedral/anhedral angle, or a neutral wings-extended position.
This extensive wing orientation control, enacted by the four actuators manipulating anhedral/dihedral and wing pitch, i.e. angle of attack, for the two wings independently, will be essential for stability for tailless gliding.
A practical control architecture will need to be studied in future work; a possible allocation is given by the following (see Figure~\ref{fig:glidectrl}):
\begin{itemize}
    \item Roll: Differential wing tilt between the left and right propulsion unit, which effectively differentially changes the incidence angle and with it the angle of attack on both wings, resulting in a differential in lift vector magnitude. Depending on the wing design, this might also result in a small yaw moment.
    \item Pitch: Symmetric wing tilt of the left and right propulsion unit, which effectively changes the incidence angle and with it the angle of attack on both wings, thus decreasing or increasing the lift vector magnitude.
    \item Yaw: This control axis cannot be straightforwardly implemented without roll cross-coupling issues; however, it might be possible to utilise the independent anhedral/dihedral-angle control of both wings to separate roll and yaw control moments.
\end{itemize}

In order to successfully implement the multimodality and omnidirectional flight capabilities of this framework, the following overarching requirements have to be met: For one, due to the individually propelled dual-wing architecture, the employed flapping mechanism needs to generate a symmetric motion profile for efficient flight and control. This balances the forces generated during upstroke and downstroke in the hover mode. Additionally, a wide flapping amplitude needs to be produced by this mechanism to allow for maximised lift generation.
This is furthermore required to exploit the lift-enhancing clap-and-fling effect with only two wings rather than four, compared to the popular X-Wing configuration that utilises this effect.
Lastly, accurate wing position tracking needs to be implemented to be able to synchronise the independent wings, guaranteeing stability and enabling wing anhedral/dihedral control during the glide mode.

\subsection{Hybrid Scotch-Yoke Flapping Mechanism}\label{flappingmechanism}
The flapping mechanism is the most important component in an FWMAV, as it dictates the wing’s motion and velocity profile during the upstroke, downstroke, and rotation phases, as well as the overall efficiency of the aerial robot. Its main objective is to convert the brushless DC motor’s rotating motion into the oscillating motion of a wing’s leading edge in an energy-efficient manner. For our framework, a symmetric and wide-angle motion profile is required as the output of the flapping mechanism.

One of the most common flapping mechanism designs is linkage-based, such as the four-bar linkage designs employed by Karasek et al.~\cite{Karasek2018}, Nguyen and Chan~\cite{Nguyen2018}, and Wu et al.~\cite{Wu2026}. However, these produce small flapping amplitudes generally under 90º, which are only suitable for X-Wing designs. To achieve larger flapping amplitudes, for two-winged designs, two stages of four-bar linkages have been tested, e.g. by Keenon et al.~\cite{Keennon2012}, who found that they feature an asymmetric motion- and thus force-generation profile leading to mechanical wear and undesired vibrations in the airframe.
Hybrid mechanisms such as the slider-crank mechanism in combination with a four-bar linkage, utilised in the two-winged Colibri~\cite{Roshanbin2017}, produced a nearly symmetric motion profile with a ~140º flapping amplitude.
String-based hybrid mechanisms can produce large flapping amplitudes, with the Nano Hummingbird~\cite{Keennon2012} featuring a 200º amplitude or the KU-Beetle~\cite{Phan2017} reaching 190º. They are, however, according to Preumont et al.~\cite{Preumont2021}, difficult to tune and sensitive to tolerances, potentially leading to asymmetries and increased friction. In their design~\cite{Preumont2021}, a gear mechanism was successfully used to replace the capabilities of the string-based mechanism.

\begin{figure}[htb]
    \centering
    \includegraphics[width=\linewidth]{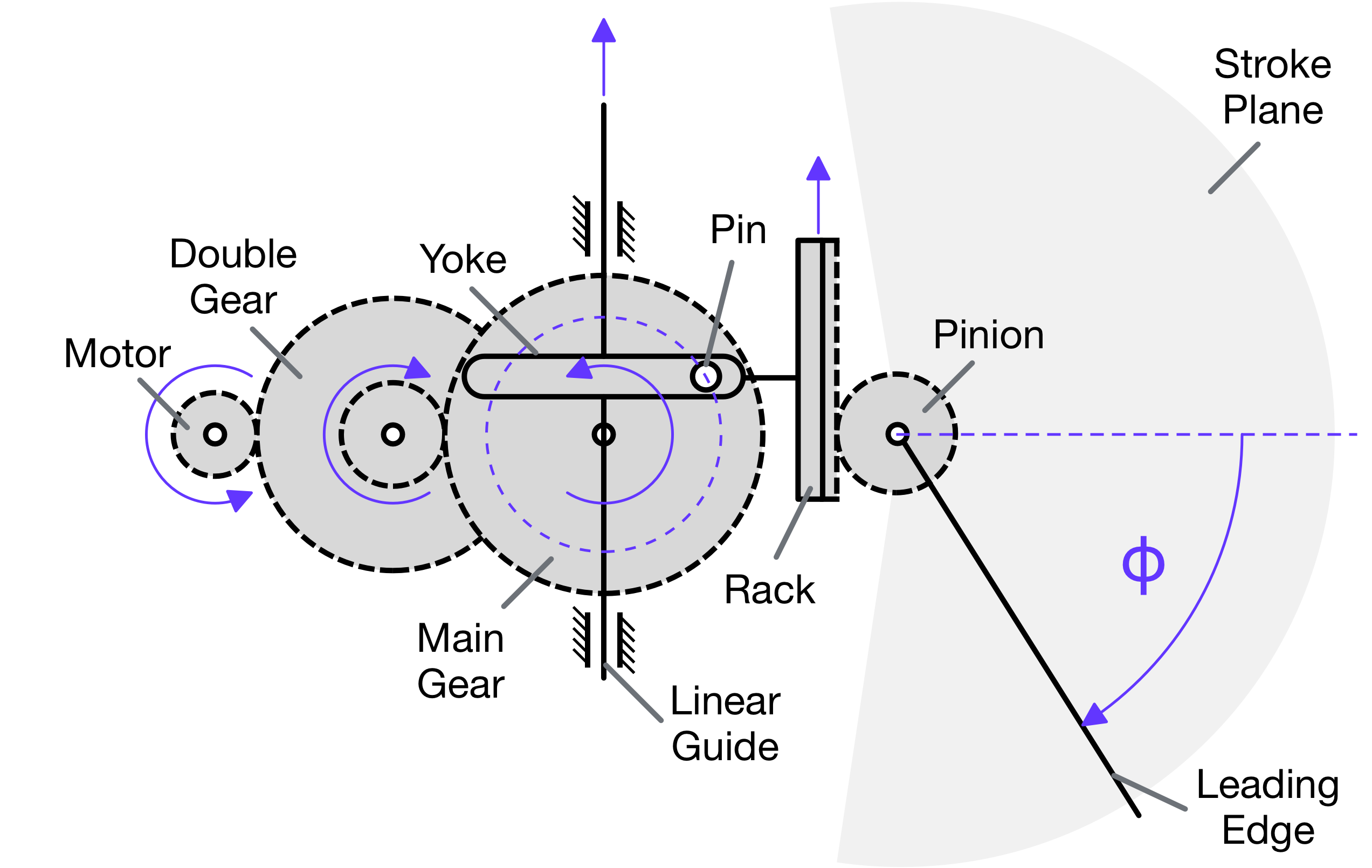}
    \caption{Functional diagram of the propulsion unit, showing the interaction between the motor, the reduction gear train, and the hybrid Scotch-yoke flapping mechanism. The output is represented by the flapping angle $\phi$, which describes the angular position of the leading edge during its symmetric oscillation in the stroke plane. For simplicity, the architecture shown has been flattened; the rack-and-pinion assembly, and therefore the stroke plane, are directly attached, in the physical prototype, at the linear guide and rotated by 90º around its axis (see Figure~\ref{fig:propunit}). The leading edge and resulting stroke plane are shown not to scale.}
    \label{fig:mechanism}
\end{figure}

For our requirements, we propose a hybrid Scotch-yoke-based mechanism, as the Scotch-yoke generates a harmonic sinusoidal motion profile from a constant-speed rotating motion and features no linkages compared to the other symmetric slider-crank mechanism.
The Scotch-yoke mechanism consists of a pin mounted to a motor-driven crank, which moves through and drives a slotted yoke that is attached to a linear guide. The linear guide exhibits the reciprocating output motion and is mounted perpendicular to the orientation of the slot. To convert the resulting harmonic linear motion of the first stage into a wide-angle flapping motion, a second stage consisting of a rack-and-pinion linear actuator is used. The linear gear (i.e. the rack) is fixed in parallel to the oscillating linear guide and used to drive a pinion gear, which is bearing-mounted to the transmission frame. A functional diagram of the complete propulsion unit assembly with its flapping mechanism is shown in Figure~\ref{fig:mechanism}.

Our mechanism allows for a scalable transmission design with easily changeable and controllable parameters, making it suitable for a wide range of flapping-wing MAVs.
This is due to the second gear stage, which can scale well with changing torque demands and the simplicity in flapping amplitude generation, compared, for example, to a double four-bar linkage, where each bar needs to be precisely designed, optimised, and machined to achieve a desired output, as seen in the work by Deng et al.~\cite{Deng2020}.
Furthermore, combining the Scotch-yoke with the rack-and-pinion gear stage allows for a simple way to achieve large, arbitrarily defined flapping amplitudes to utilise the highly efficient clap-and-fling capability for the FWMAV. These large amplitudes are needed to achieve near contact of the wing leading edges after each stroke. In a two-winged configuration, this requires the wing tips to nearly meet at each end of the flapping stroke; therefore, the maximum needed flapping angle $\phi_{max}$ can be expressed as a function of the wing's semi-span $s$ and the design-driven centre-of-rotation offset $D_y$ along the robot body's $Y_B$ axis (i.e. half the distance between both flapping axes, see Figure~\ref{fig:fwcomponents}):
\begin{equation}
    \phi_{max} = 2 \arcsin \!\left( \frac{D_y}{s} \right) + \pi
\end{equation}

When a feasible pinion diameter $d_{pinion}$ is chosen, depending on torque-demands, resulting in a specific gear-teeth count and module, the minimum size of the rack $l_{rack}$ and therefore the minimum length of the slotted yoke $l_{yoke}$ (i.e. double the eccentricity of the pin $e_{pin}$) directly dictates the maximum flapping angle $\phi_{max}$:
\begin{equation}
    e_{pin} = \frac{1}{2}  l_{yoke } = \frac{1}{2}  l_{rack} = \frac{1}{4} \phi_{max} d_{pinion}
\end{equation}

Changing only one design parameter of the Scotch-yoke part of the mechanism, i.e. $e_{pin}$, allows for an easy alteration of the crucial flapping angle and a straightforward way to manufacture different FWMAV configurations. 
These design features allow the flapping mechanism to easily scale to the required FWMAV sizing.

The full propulsion unit consists of a small brushless DC motor, a reduction gear train, the flapping mechanism, and a wing. For prototyping purposes, a larger 3D-printed PLA-based frame was developed, which allowed quick alterations of the reduction gear stage and the attachment of different motors.
Although the framework and the developed flapping mechanism aim to be scalable and thus applicable for varying sizes of FWMAVs, a target size of around 30\,g was chosen for the initial prototype presented in this paper.
This size class was chosen as it features most tailless FWMAVs, enabling a comparison in capabilities. For example, the previously mentioned X-wing NUS-Roboticbird~\cite{Nguyen2018} weighed 27\,g, featuring two 2.5\,g BLDC motors, while the Colibri~\cite{Roshanbin2017} weighed 22\,g with a single 4.9\,g DC motor.
Overall, robot size and lift-production capabilities are largely dependent on the motor size chosen for the propulsion unit
To position our prototype in this size class, we therefore chose the 1S AP03 7500\,KV brushless DC motor weighing 3.1\,g.
The gears used in the transmission are module 0.5 plastic toy gears, which are combined with plastic IGUS sleeve bearings paired with shafts made from brass tubes.
The yoke of the flapping mechanism is made of 3D-printed PC for enhanced strength and features small brass tubes acting as railings for the pin to slide against. The pin itself is mounted to the last spur gear in the reduction gear train and made from a brass tube, around which a small nylon sleeve rotates to reduce friction with the yoke railings. The yoke is fixed to the rack and the linear guide, which is made from a brass tube sliding through sleeve bearings that are mounted to the frame of the transmission. The pinion that meshes with the rack is mounted on a brass tube shaft, supported via two sleeve bearings on each side, which are fixed to the frame.
Attached to the pinion is the wing joint, which contains a passive wing-pitch mechanism. This mechanism comprises a revolute joint which allows the wing to passively rotate around the spanwise axis of its leading edge.
This rotation is symmetrically limited on two sides through mechanical endstops enforcing a defined maximum wing pitch angle $\theta_{max}$ during the translation phase of the flapping cycle while allowing free rotation of the wing during stroke reversal.

A similar design has been proposed by Keenon et al.~\cite{Keennon2012} with an active endstop modulation system enforced by two arms stopping the root chord, while Coleman et al.~\cite{Coleman2017} introduced a flexible shim between the leading edge and root chord to allow for passive wing rotation.
This allows additional wing design freedom as opposed to the most common solution that consists of a fixed leading edge and a fixed root chord with a specific degree of slack in the wing membrane enabling passive wing rotation and resulting in a maximum wing pitch angle. 
X-wing FWMAVs (e.g.~\cite{Nguyen2018,Karasek2018,Wu2026}) utilise this design as a practical measure to combine two wings with a single root chord, while most two-winged designs need this to exert control through active root-chord manipulation (e.g.~\cite{Keennon2012,Phan2017,Roshanbin2017,Preumont2021}).

In the current prototype, an aluminium tube is paired with the extension of the leading edge carbon fibre rod to serve as the revolute joint.
Extrusions fixed to the leading edge and root chord connector act as physical endstops when they come in contact with a cap that is screwed onto the rotary joint, which also restricts any axial movement of the leading edge.
Altering the cap and endstop geometries results in precise maximum wing pitch angle limits.
The optimal pitch limit for our propulsion unit was experimentally determined in Section~\ref{pitchlimit}.

\begin{figure}[htb]
    \centering
    \includegraphics[width=0.8\linewidth]{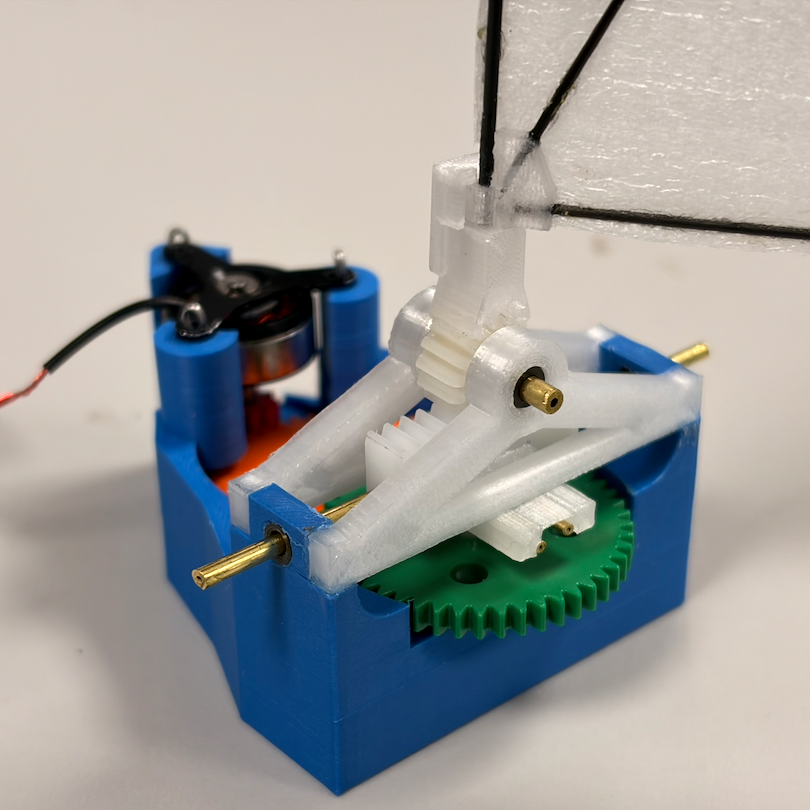}
    \caption{Assembly of the propulsion unit for preliminary testing, containing the motor, reduction gear train, flapping mechanism, and a preliminary wing attached via the passive wing rotation mechanism.}
    \label{fig:propunit}
\end{figure}

The complete prototype assembly of the propulsion unit for preliminary performance testing can be seen in Figure~\ref{fig:propunit}.

For simplicity and preliminary testing, an unoptimised rectangular wing was designed for the propulsion unit.
Common wing designs (e.g.~\cite{Keennon2012,Coleman2017,Roshanbin2017,Nguyen2018,Guo2024}) feature a sail-like construction comprising a leading edge and root chord made from carbon-fibre rods spanning a thin membrane, with various planforms, made from Mylar~\cite{Nguyen2018}, polyester film~\cite{Roshanbin2017,Wu2024,Wu2026} or foam~\cite{Coleman2017} on which thinner carbon fibre stiffeners are placed at various positions. 
The 22\,g Colibri by Roshanbin et al.~\cite{Roshanbin2017} features a 90$\times$25\,mm wing weighing 480mg, while the wing by Coleman et al.~\cite{Coleman2017} weighed 850mg with a similar aspect ratio of 3.7 and a 140$\times$38\,mm wing size for a 62\,g robot. The 27\,g X-wing design of Nguyen and Chan~\cite{Nguyen2018} featured a wing with a 100x60mm size weighing 400\,mg.
Our wing features a simplified rectangular design and serves as a stand-in to test the performance of the flapping mechanism and to gauge the general capabilities of our framework. However, future work will focus on optimising the wing design and tailoring it to our configuration.
We chose an aspect ratio of $AR \approx 2$ with a wing size of 130$\times$60\,mm following the above-mentioned examples representing a comparable size class. The wing features a leading edge made of a 1\,mm diameter carbon-fibre rod and a 0.8\,mm diameter carbon-fibre root chord to provide sufficient rigidity. Additionally, a diagonal 0.8\,mm diameter carbon-fibre rod acts as a supporting wing stiffener. These three carbon-fibre rods are joined at the intersection of the root chord and leading edge via a small 3D-printed connector made of PC. All parts were glued together onto a 1\,mm thick low-density polyethylene foam membrane, resulting in a mass of 650\,mg.

For the transmission to run efficiently and effectively, the speed output demands of the flapping mechanism at a given supply voltage need to match the motor torque required. This is achieved with a reduction gear train, letting the motor run at an efficient speed while increasing the mechanism output torque to the required level. Too low reduction ratios will result in the motor running at an inefficient operating point, not being able to generate sufficient torque. A transmission with a too large reduction rate, however, will generate sufficient torque but won't produce the desired output speeds.

Because the motor performance curves and ideal operating points are often not known for common off-the-shelf BLDC motors, an experimental approach is chosen to systematically test different gear trains for their propulsion efficiency.
Our selected motor runs at a maximum no-load speed of 27,750\,rpm at a voltage of 3.7\,V, resulting in a theoretical no-load flapping frequency of 462.5\,Hz, without a reduction gear train. 
From the literature, the first indication of reduction-ratio sizing for BLDC motors can be gauged: Coleman et al.~\cite{Coleman2017} used a 9.3$\times$ reduction, and Guo et al.~\cite{Guo2024} used a 10$\times$ reduction. Karasek et al.~\cite{Karasek2018} used a reduction of 21.3$\times$, while Nguyen and Chan~\cite{Nguyen2018} utilised a similar motor to ours with a 20$\times$ reduction ratio.

\begin{figure}[htb]
    \centering
    \begin{subfigure}[t]{\linewidth}
        \centering
        \includegraphics[width=0.7\linewidth]{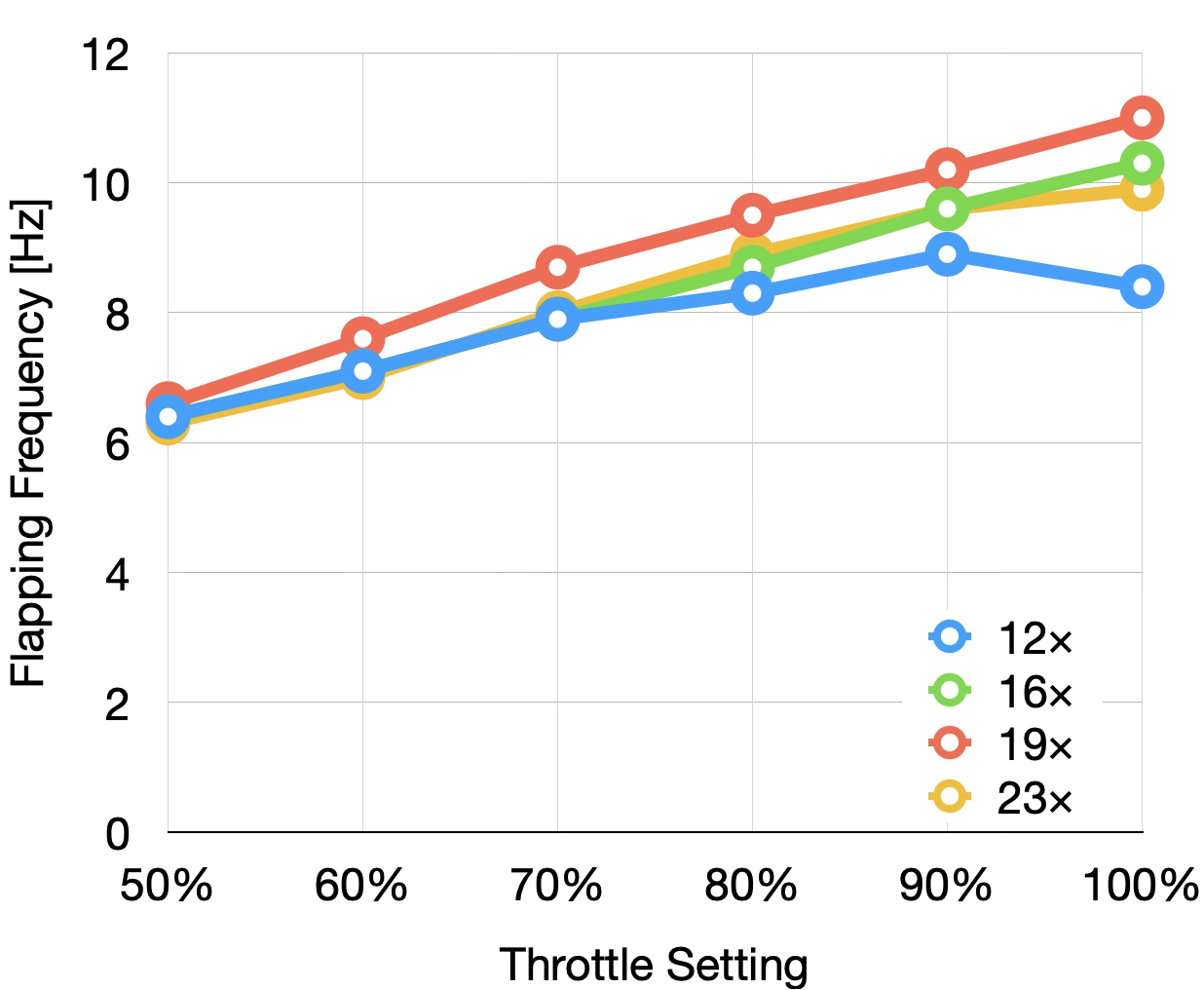}
        \caption{Resulting flapping frequency over the throttle range for different reduction gear stages.}
        \label{fig:gearfrequency}
    \end{subfigure}

    \vspace{1em}

    \begin{subfigure}[t]{\linewidth}
        \centering
        \includegraphics[width=0.7\linewidth]{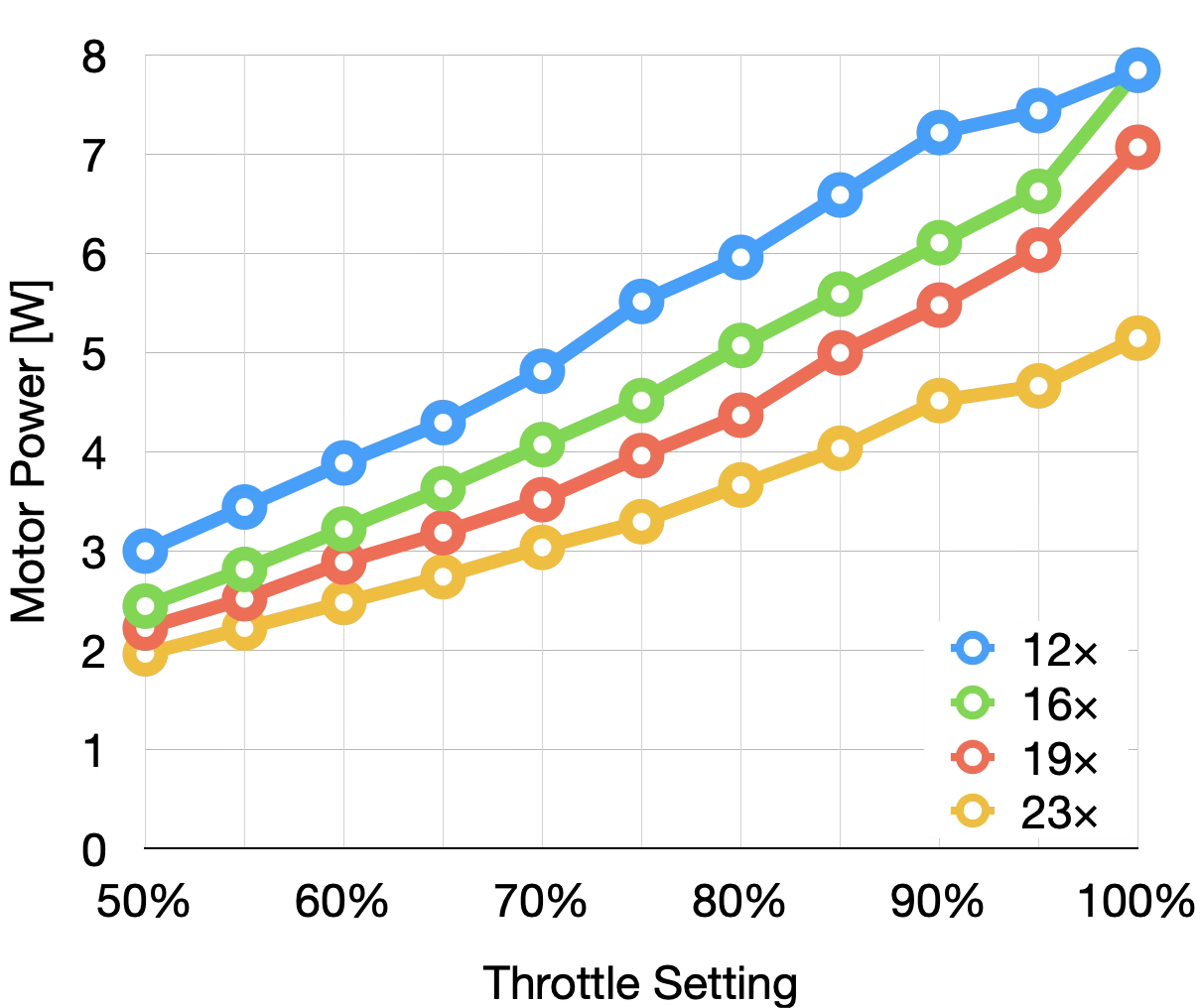}
        \caption{Motor power drawn over the throttle range for different reduction gear stages.}
        \label{fig:gearpower}
    \end{subfigure}
    \caption{Experimental results of testing different reduction gear ratios for the flapping mechanism to gauge transmission efficiency and lift production.}
    \label{fig:gearratio}
\end{figure}

In order to systematically test a suitable reduction ratio, a prototyping assembly of our propulsion unit is set up, which allows the installation of different reduction gear trains. This is achieved via stages of double spur gears connecting the driving gear of the motor and the flapping mechanism.
By exchanging the sizes of the double spur gears, four reduction ratios were produced and tested, i.e. 12$\times$, 16$\times$, 19$\times$, 23$\times$.
To evaluate the transmission’s efficiency, motor power consumption was recorded in addition to the resulting flapping frequency using a simple rectangular wing at varying throttle settings, see Figure~\ref{fig:gearratio}.
The flapping frequency can hereby be used as a proxy for lift generation, as it has been shown to correlate well \cite{Nguyen2017, Nguyen2018, Preumont2021}.

Resulting from these tests, as seen in Figure~\ref{fig:gearfrequency}, it is evident that the 19$\times$ reduction ratio performed best, leading to higher flapping frequencies and therefore higher lift over the throttle range compared to the other gear trains.
For the motor performance, shown in Figure~\ref{fig:gearpower}, it follows that the larger the reduction ratios are, the lower the power consumption is at each throttle setting. Therefore, the 23$\times$ reduction ratio would be best performing from a pure efficiency standpoint; however, the 19$\times$ reduction is still chosen for the transmission design as it produces significantly more lift than the 16$\times$ version.

To generalise these results, a 19x reduction ratio in combination with our motor produces an ideal no-load output of ca. 24\,Hz, which under load results in a 11\,Hz output, i.e. 46\% of the maximum no-load speed.
Overall, this result is in line with the gear reductions chosen by Karasek et al.~\cite{Karasek2018}, and by Nguyen and Chan~\cite{Nguyen2018} for their similar speed and size BLDC motor.

\subsection{Wing Position Tracking}
Another important component to realise the proposed framework is the sensor that enables accurate tracking of the angular orientation $\phi$ of the wing’s leading edge. This is needed to enable the synchronised operation of the independently actuated two-winged design. Additionally, it is needed to provide feedback for the dihedral/anhedral angle control of the wings in the glide-mode configuration. In this mode, the stroke plane is tilted forward, and the bidirectional BLDC motor acts as a closed-loop servo actively changing the dihedral/anhedral angle of each wing. 
A dihedral angle on both wings, i.e. flying with wings inclined, provides enhanced roll stability. This is frequently utilised in general aviation and observed in many slow gliding birds \cite{Thomas2001}. An anhedral angle of both wings, i.e. flying with wings depressed, grants enhanced roll manoeuvrability, especially useful during fast glides as observed in agile swifts \cite{Thomas2001}.

A commonly used approach to motor position control is realised via a motor encoder and the calculation of the theoretically resulting position of the flapping mechanism and gear train, e.g. by Guo et al.~\cite{Guo2024}.
However, we propose a direct leading-edge sensor implementation that provides enhanced accuracy by ignoring the errors induced by play in the transmission and flapping mechanism, as well as by general manufacturing imprecision.
For this, a two-axis Hall-effect sensor measuring the angular orientation of the leading edge at its centre of rotation presents a promising solution, as it is compact and lightweight, consumes little energy and enables contactless sensing, therefore preserving the mechanism’s performance. We designed a custom sensor board using the Infineon TLE493D-A2B6 digital three-axis Hall-effect sensor, which will be used in its two-axis configuration to determine the angular orientation of a small cylindrical N45 magnet. This magnet is mounted on the shaft of the pinion that holds the wing and oriented with its longitudinal axis parallel to the leading edge and its centre of gravity coinciding with the centre of rotation of the wing.
The contactless Hall effect sensor is fixed to the transmission frame, hovering at a distance of 2\,mm above the magnet’s centre of rotation. It measures the magnetic field strength from the magnet along the X-axis and Y-axis, which is communicated over the I2C protocol with the flight controller, or in this prototyping setup, an Arduino Uno microcontroller. There, the actual angular orientation of the magnet and therefore the leading edge is calculated via $\phi = atan2(B_y,B_x)$ and output as a positive or negative angular value with 0º at the middle position in the flapping cycle. For future work, this sensor board will be implemented in a compact and lightweight integrated PCB.

After integrating the sensor board on the transmission frame and implementing the optimal reduction ratio, the propulsion unit underwent further improvements to establish a reliable prototype to commence extended testing of its capabilities.

\section{Multibody and Aerodynamics Model}\label{multibody}
\begin{figure}[htb]
    \centering
    \includegraphics[width=\linewidth]{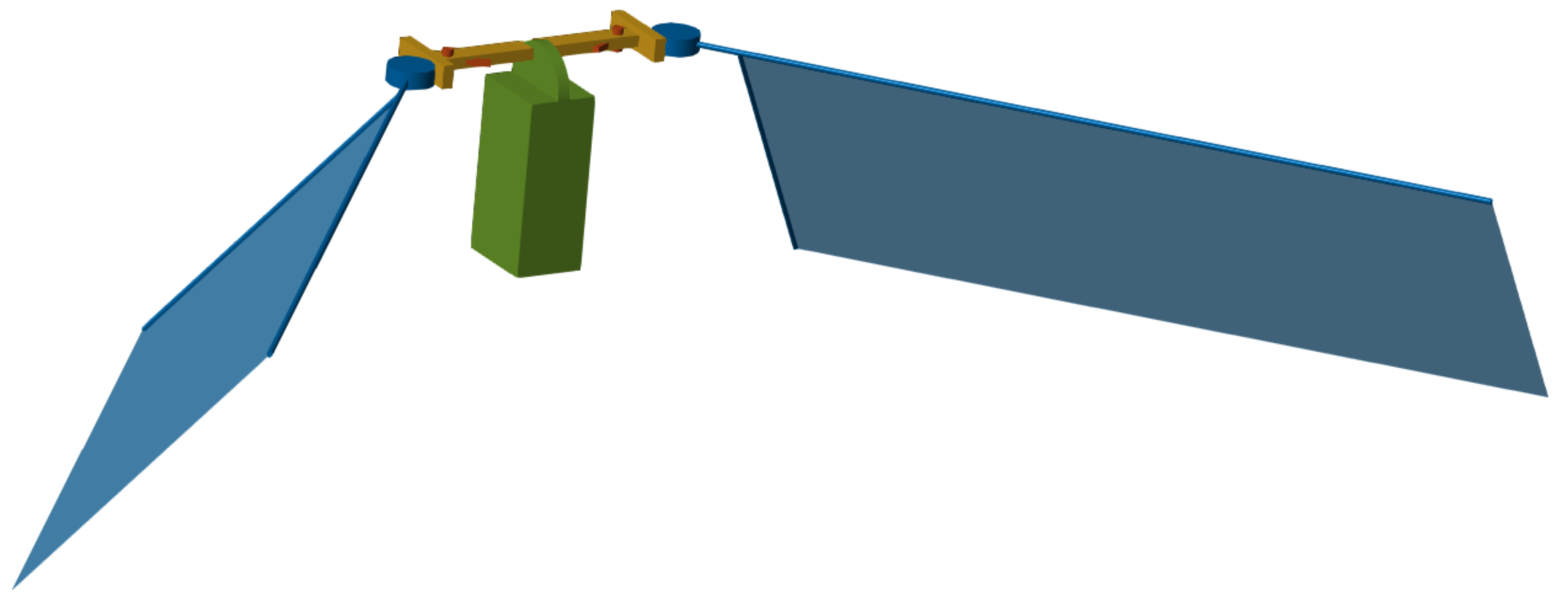}
    \caption{3D view of the Simscape Multibody physics model, showing both flapping mechanisms (see also Figure~\ref{fig:mechanism}) and wings in a flapping motion, as well as a simplified rectangular body representing the robot's remaining components and mass with a specified centre of gravity offset.}
    \label{fig:3dmultibody}
\end{figure}

Parallel to the physical design, a multibody physics simulation of the flapping mechanism and aerial robot was established to aid the design process and give initial estimates on performance.
During the flapping mechanism design, MATLAB Simulink Simscape Multibody allowed quick iteration on potential concepts and gave insight into joint and constraint configuration, with the ability to display the produced motion profile.
With the hybrid Scotch-yoke mechanism described in Section~\ref{flappingmechanism} modelled, a parametric model of the full robot configuration was established. It features two flapping mechanisms actuated by two ideal constant-speed motors powering two rectangular wings. 
These propulsion units are jointed to the robot’s frame by two actuated revolute joints, representing the tilting servo-motors.
The robot’s remaining components, such as avionics, battery, and frame structure, were simplified into a rectangular body and a set centre-of-mass offset to the propulsion unit attachment point. The resulting 3D multibody model of the robot's assembly can be seen in Figure~\ref{fig:3dmultibody}.

In order to estimate the robot’s flight behaviour in the multibody simulation environment, an aerodynamics model was established to calculate the forces generated by the flapping wings.
For this, a quasi-steady model utilising the blade element theory (BET) was implemented, adapted from the insect-wing model proposed by Dickson et al.~\cite{Dickson2008}
The flapping wings were modelled to consist of a stiff leading edge with an attached root chord rod that spanned a stiff membrane. They were attached to the oscillating flapping mechanism output by the passive wing-pitch mechanism detailed previously, which consists of a limited revolute joint, allowing the wings to rotate around their spanwise leading-edge axis.
This results in a defined pitch angle $\theta_{max}$ during the wing’s translation phase and a passive rotation at the end of each stroke.
To more accurately predict the aerodynamic forces, the wing’s current pitch angle $\theta$ was included in the BET calculation. The output of this aerodynamic model for each wing individually was implemented as an external force acting on the aerodynamic centre of the wing, for simplification assumed to be located at a fixed point at the quarter chord length and the three-quarter span length.

To calculate the steady-state forces resulting from lift and drag in the wing’s reference frame, first, the angle of attack $\alpha$ for a blade element is calculated, taking into account the pitch angle $\theta$ of the wing and the freestream velocity $V$:

\begin{equation}
    \alpha(r) = \arctan\!\left(\frac{ \cos \theta \, \dot\phi r - \cos \theta \, V_j + \sin \theta \, V_i }{ -\sin \theta \, \dot\phi r + \sin \theta \, V_j + \cos \theta \, V_i }\right)
\end{equation}

\begin{figure}[htb]
    \centering
    \includegraphics[width=0.8\linewidth]{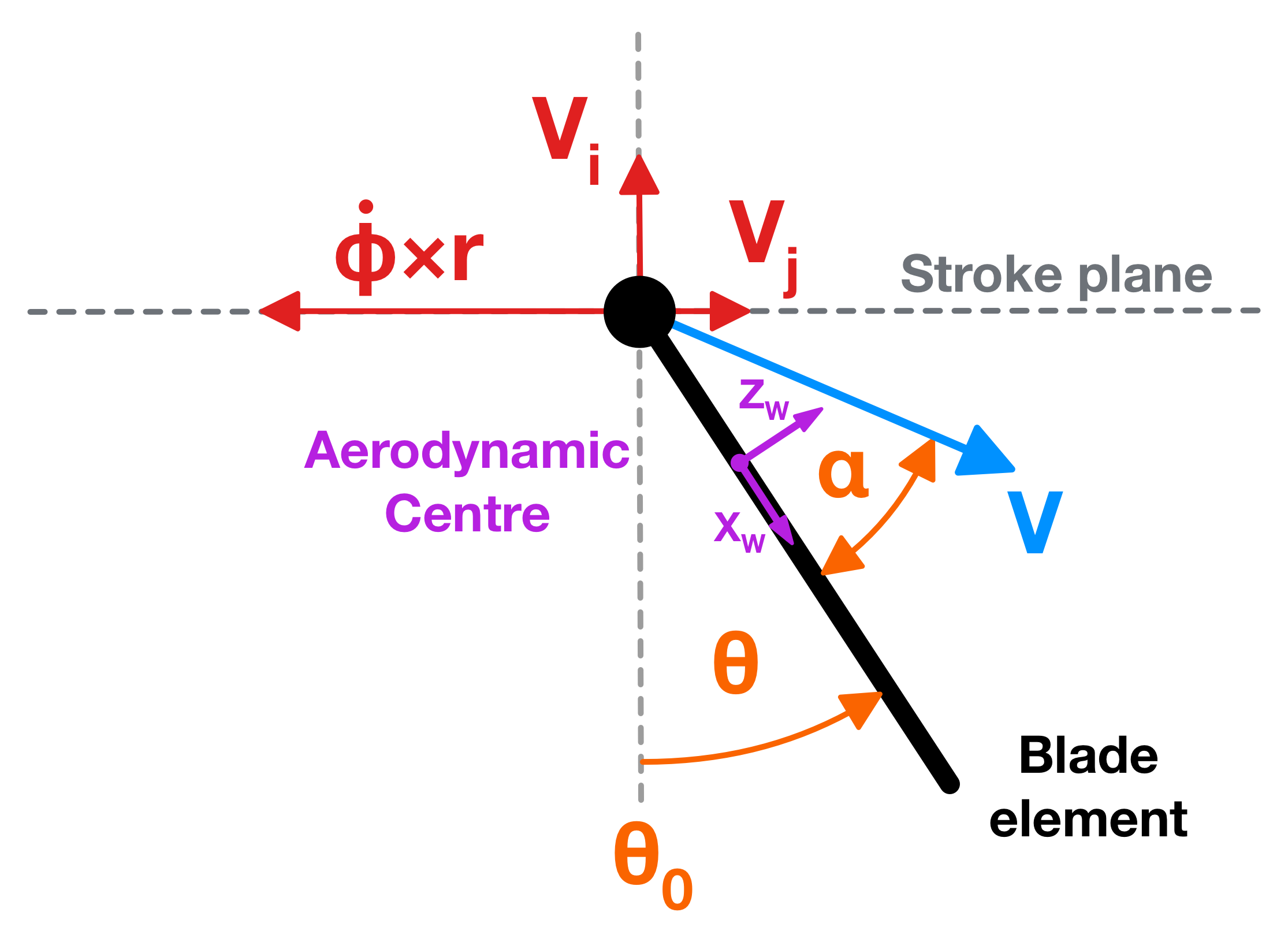}
    \caption{Side view of a blade element showing the freestream velocity components ($V_i$,$V_j$, and $\dot\phi*r$), the angle of attack~$\alpha$, the pitch angle~$\theta$, and the wing's reference frame for the aerodynamic centre.}
    \label{fig:bladelement}
\end{figure}

The freestream velocity $V$ is made up of the freestream velocity components $V_i$,$V_j$, and $\dot\phi*r$:
$V_i$ represents the velocity component normal to the stroke plane, while $V_j$ represents the component acting tangential to the stroke plane.
The angular velocity $\dot\phi$, i.e. the derivative of the flapping angle $\phi$ of the leading edge revolving around the flapping axis, multiplied by the radial distance $r$, represents the other free stream velocity component of the flapping wing in its stroke plane.
In Figure~\ref{fig:bladelement}, a chord view of the wing displays the orientation and definition of the frames and angles of the blade element.

With this, the lift force $\delta F_{lift}$ and drag force $\delta F_{drag}$ equations are formulated for the blade element. Here, $C_l$ is the aerofoil's lift coefficient, $C_d$ is the aerofoil's drag coefficient, and $c$ is its chord length for the rectangular wing.
\begin{equation}
    \delta F_{lift}(r) = \frac{1}{2} \rho C_l(r) c \left( V_i^2 + \dot\phi^2 r^2 - 2 \dot\phi r V_j + V_j^2 \right) \text{sgn}(\alpha(r))
\end{equation}
\begin{equation}
    \delta F_{drag}(r) = \frac{1}{2} \rho C_d(r) c \left( V_i^2 + \dot\phi^2 r^2 - 2 \dot\phi r V_j + V_j^2 \right)
\end{equation}

Following a frequent approach taken in the literature, the lift and drag coefficients come from an experiment done by Dickinson et al.~\cite{Dickinson1999} approximating a fruit fly wing with a Reynolds number of $Re =136$.
They obtained the following simple harmonic functions for the lift and drag curve~\cite{Dickinson1999} with $\alpha$ in radians\footnote{NB: While our lift and drag coefficient curves are identical to those in the underlying paper, there is a discrepancy in the phase-shift parameter in both equations, as the paper appears to have omitted a degree-to-radians conversion.}:
\begin{equation}
    C_l = 0.225 + 1.58 \sin(2.13\alpha -0.1257) 
\end{equation}
\begin{equation}
    C_d = 1.92 - 1.55\cos(2.04\alpha-0.1714) 
\end{equation}

Our design, with its preliminary sizing, would operate at a larger Reynolds number of $Re\approx 3.8\times10^4$ for a flapping frequency of 10\,Hz; nonetheless, these coefficients represent a useful approximation, as more suitable data is not available.
A similar approximation for these coefficients, despite Reynolds number differences, has been made, for example, by Armanini et al.~\cite{Armanini2019}.

Transforming the lift and drag force of the blade element into the wing’s reference frame (shown in Figure~\ref{fig:bladelement} denoted with a subscript $W$) and integrating over the wing span, we have the resultant tangential steady-state force $F_{s,x}$ and the resultant normal steady-state force $F_{s,z}$:
\begin{equation}
    F_{s,x} = \int_{r_0}^{R} \delta F_{drag}(r) \cos \alpha(r) - \delta F_{lift}(r) \sin \alpha(r) \, dr
\end{equation}
\begin{equation}
    F_{s,z} = \int_{r_0}^{R} \delta F_{lift}(r) \cos \alpha(r) + \delta F_{drag}(r) \sin \alpha(r) \, dr
\end{equation}

Following Chin and Lentink~\cite{Chin2016}, the additionally considered aerodynamic force components for the quasi-steady BET model are the rotational force $F_{rot}$ and the added-mass force $F_{add}$. The rotational force incorporates the rotational coefficient $C_r=1.55$, as utilised by the quasi-steady BET model of Sane and Dickinson~\cite{Sane2002}, and the pitch angular velocity $\dot\theta$.
\begin{equation}
    F_{rot} = \int_{r_0}^{R} C_r \rho \dot{\theta} c^2 \sqrt{ V_i^2 + \dot\phi^2 r^2 - 2 \dot\phi r V_j + V_j^2 } \, dr
\end{equation}

In the added-mass force equation, derived from inviscid flow theory~\cite{Chin2016}, where the wing is an assumed flat plate of infinitesimal thickness \cite{Dickson2008}, both square roots represent the freestream acceleration $\dot V$.
\begin{equation}
    \begin{split}
    F_{add} = \int_{r_0}^{R} \frac{1}{4} \rho \pi c^2 \sqrt{ \dot{V_i^2} + \ddot\phi^2 r^2 - 2 \ddot\phi r \dot{V_j} + \dot{V_j^2}} \sin \alpha(r)+ \\+ \sqrt{ \dot{V_i^2} + \ddot\phi^2 r^2 - 2 \ddot\phi r \dot{V_j} + \dot{V_j^2}} \alpha(r) \cos \alpha(r) \, dr
    \end{split}
\end{equation}
With all individual force components calculated, the resultant aerodynamic forces are then applied in the wings's reference frame, where $F_{A,x}$ acts chordwise in $X_W$-direction, $F_{A,y}$ acts spanwise in $Y_W$-direction, and $F_{A,z}$ acts normal to the wing's surface in $Z_W$-direction (see Figure~\ref{fig:bladelement}). To note, the $Y_W$-component has been neglected in this model, due to the two-dimensional nature of the blade element theory.
\begin{align}
    F_{A,x} &= F_{s,x}\nonumber\\
    F_{A,y} &= 0\\
    F_{A,z} &= F_{s,z}+F_{rot}+F_{add}\nonumber
\end{align}

During the simulation, the individual force components are numerically integrated over the wing's span at each solver step, as the partial analytical solution resulted in an unwieldy expression. With given settings for motor speed, i.e. flapping frequency ($f_r$,$f_l$), and a desired stroke-plane tilt angle ($\beta_r$, $\beta_l$) for the left and right wing, different flight configurations can be simulated to give an initial estimate of the robot's behaviour.

\begin{figure}[htb]
    \centering
    \includegraphics[width=0.9\linewidth]{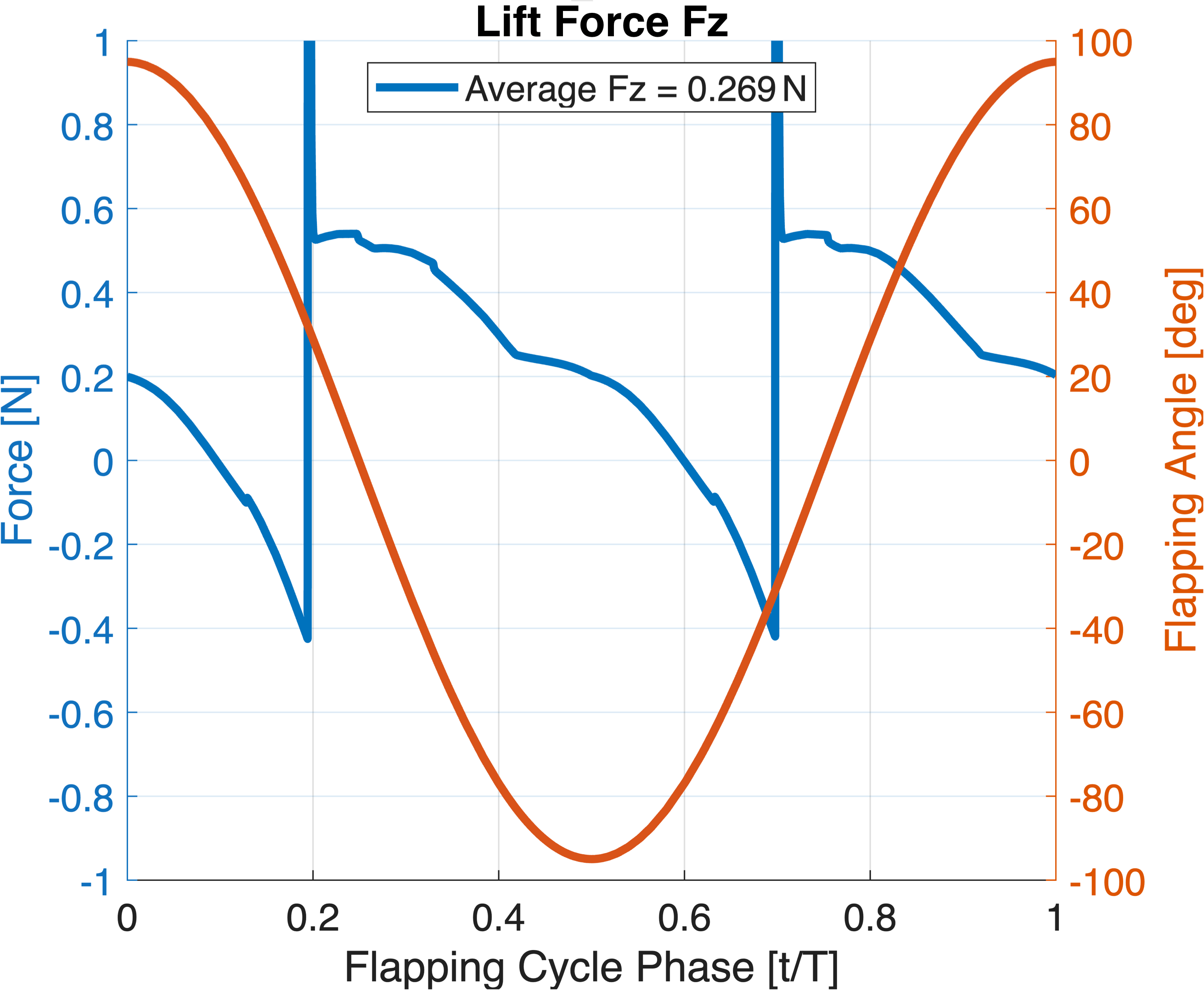}
    \caption{Simulation results for the lift force generated by the proposed single-wing propulsion unit during the flapping cycle overlaid with the trajectory of the flapping angle in orange. A cycle-averaged lift force of 0.269\,N is predicted in the hover mode for a $130\times60$\,mm wing running at 10\,Hz with an amplitude of 190º for the flapping angle.}
    \label{fig:simfz}
\end{figure}

Setting the configuration to a hover mode with a constant flapping frequency of 10\,Hz, the simulation predicts a cycle-averaged lift force of 0.269\,N generated per wing.
The simulated behaviour of the lift force within a full flapping cycle can be seen in Figure~\ref{fig:simfz}.

For this, our previously detailed exemplary wing layout with a surface area of $130\times60$\,mm was used and actuated at a constant speed, generating the predicted harmonic flapping-angle motion profile oscillating with a set amplitude of 190º, mirroring the first physical prototype.

The lift-force graph shows a symmetric lift profile for upstroke and downstroke with an initial peak in force generated after each completed wing rotation. This short peak is due to simulation inaccuracies in the wing rotation joint and the torque profile generated by the simplified constant-speed motor at the time of impact of the rotating wing with its defined pitch limit $\theta_{max}$.
In general, the highest lift forces are generated at the beginning of the translation phase of the wing, with the lowest forces generated during wing rotation at the end of each stroke. Of note, the timing of the wing rotation represents a slightly offset symmetric reversal, due to the passive wing rotation, facilitated by inertial forces and drag on the wing.

\begin{figure}[htb]
    \centering
    \includegraphics[width=\linewidth]{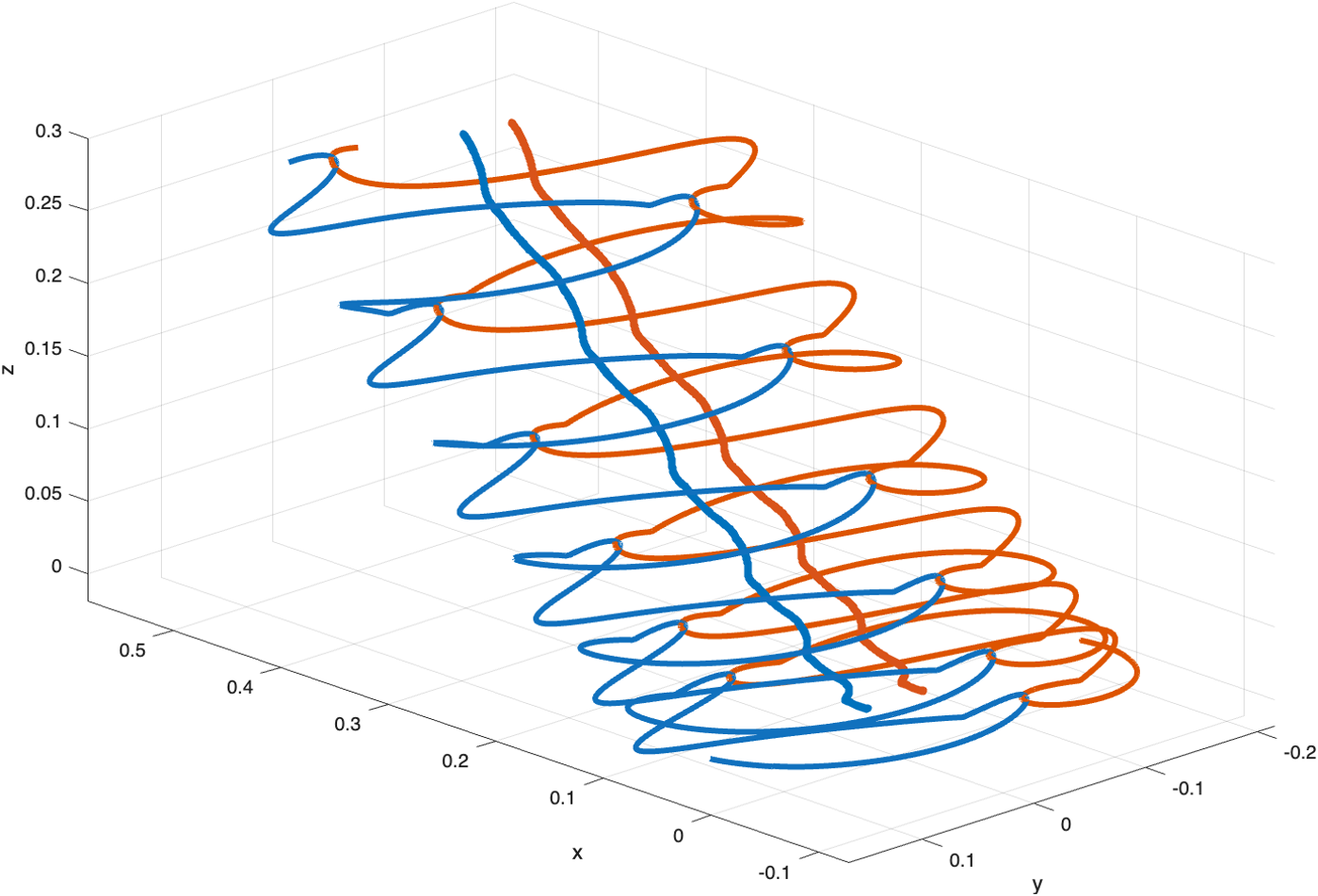}
    \caption{Wing tip paths and centre-of-rotation trajectories for the left and right wing generated from simulation data for a directional take-off flight with a small symmetric tilt angle and a constant flapping frequency of both wings.}
    \label{fig:wingtip}
\end{figure}

In addition to the observation of individual variables or forces generated, the simulation shows a 3D animation of the system’s dynamics, and a wing tip trajectory profile was plotted for further insight, see Figure~\ref{fig:wingtip}.

\section{Force/Torque Testing and Results}\label{testing}
Before manufacturing a weight- and tolerance-optimised propulsion unit for integration into a flight-capable robot, the performance of the mechanism design and the control architecture needed to be extensively tested and quantified. For this, a test rig was set up to acquire the forces, torques, wing orientation, and other performance metrics generated by the propulsion unit, such as transmission efficiency. At first, with the established setup, the passive wing-rotation mechanism was tested for optimal design parameters. With the best-performing version, a series of tests was then conducted to quantify the unit’s throttle behaviour in a hover-mode configuration to show the lift generation phases during the flapping cycle. Following the hover mode testing, the potential of the proposed wide-angle tilt mechanism was studied over the full envisaged range. These tests simulate directional flight behaviour and the control input efficacy for hovering flight. Assuming a specified robot configuration, the control moments generated were calculated from test data. Following this, the glide mode was tested with a demonstration of the active dihedral/anhedral actuation using the motor as a closed-loop servo with the leading-edge-tracking sensor. Lastly, the gliding performance under external airflow was quantified on the test rig with a range of wing tilts.

\subsection{Experimental Setup}
\begin{figure}[htb]
    \centering
    \includegraphics[width=\linewidth]{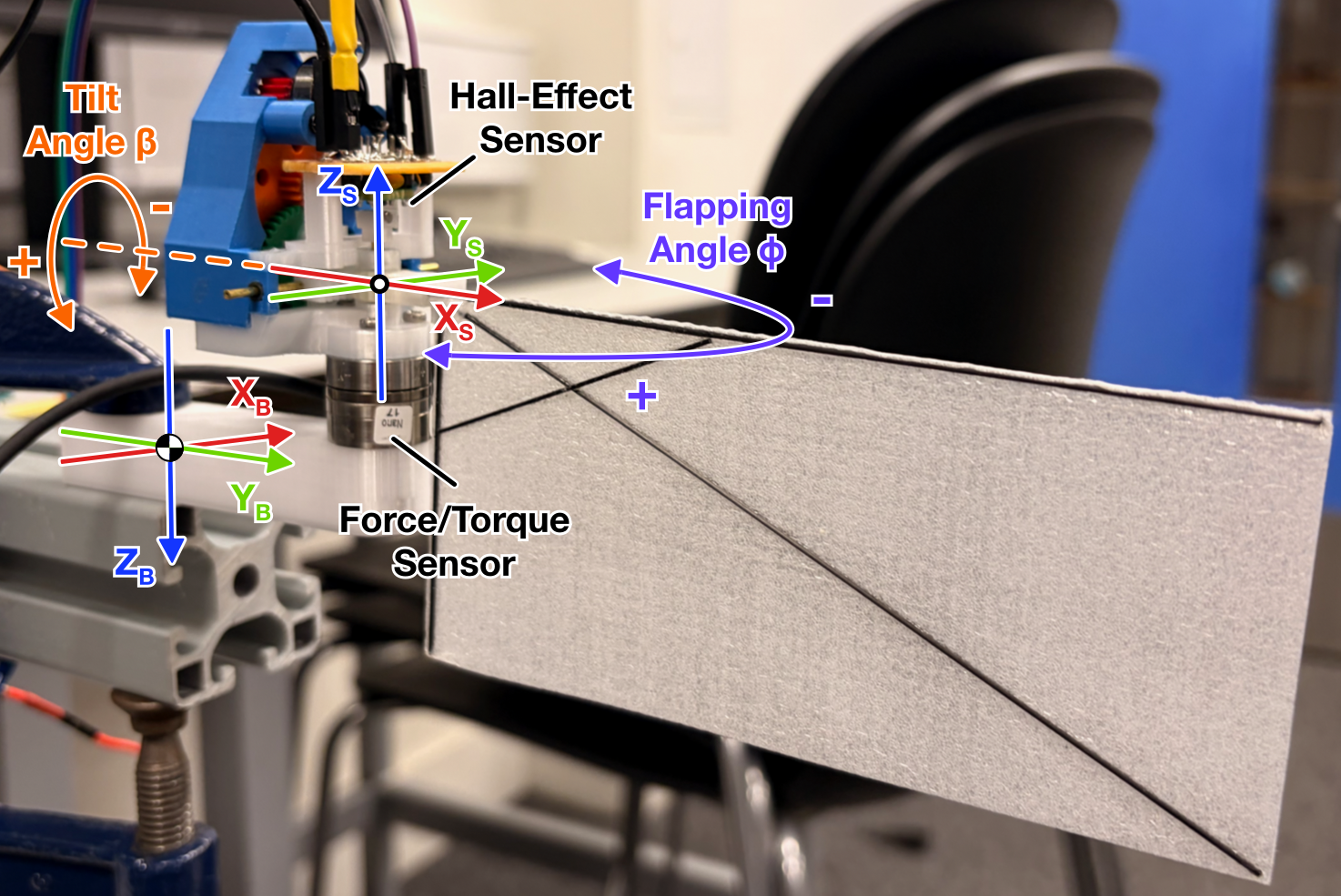}
    \caption{Detailed view of the experimental setup containing the propulsion unit prototype with integrated Hall-effect flapping-angle sensor and the force/torque sensor with its effective measurement axes, i.e. the sensor frame, denoted with subscript $S$. The location of the body-fixed frame, denoted with subscript $B$, is shown with the centre of gravity. Additionally shown is the definition of the tilting angle for the propulsion unit, and flapping angle for the wing's leading edge.}
    \label{fig:ftsetup}
\end{figure}

To accurately gauge performance metrics of the proposed propulsion unit, a test rig was built around a six-axis force/torque sensor. The transducer chosen was the commonly used ATI Nano17 Titanium, as it features a compact design and a high resolution of 0.0029\,N, ideal for small loads.
An adapter plate was designed to securely screw the propulsion unit onto the load cell, with the wing’s axis of rotation coinciding with the load cell’s $Z$-axis, see Figure~\ref{fig:ftsetup}.
The stroke plane in which the leading edge travels is parallel to the sensor’s $XY$-plane and offset by only 11\,mm.
To set the effective measurement point to coincide with the leading edge, as this point transfers the forces and torques generated to the propulsion unit's frame, a coordinate frame transformation is performed, with the resulting effective measurement axes of the sensor, i.e. the sensor frame, denoted with the subscript $S$, shown in Figure~\ref{fig:ftsetup}.

\begin{figure}[htb]
    \centering
    \includegraphics[width=0.8\linewidth]{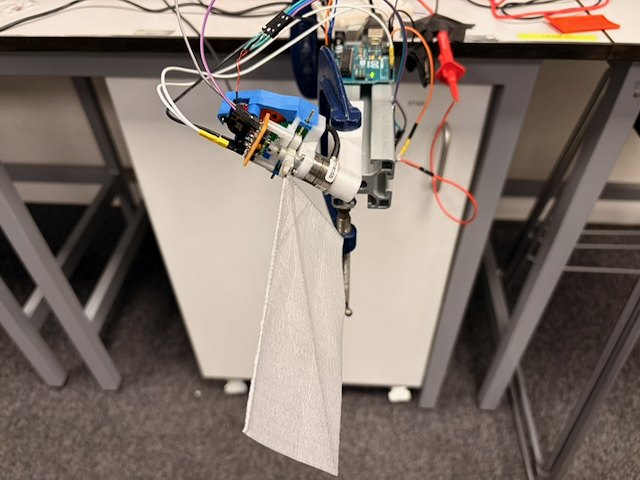}
    \caption{Orientation of the propulsion unit during testing of the tilting capabilities with a tilt angle $\beta = +60^{\circ}$.}
    \label{fig:tilt}
\end{figure}

The base of the sensor is then mounted on a small bridging piece such that the wing's root chord can move unhindered, and the propulsion unit is suspended freely to avoid any ground effect. This bridge, holding the load cell and propulsion unit, can be rotated around its longitudinal axis and fixed in a position to emulate the proposed tilting capabilities. In Figure~\ref{fig:tilt} the orientation of the propulsion unit for an exemplary tilt angle is shown.

\begin{figure}[htb]
    \centering
    \includegraphics[width=\linewidth]{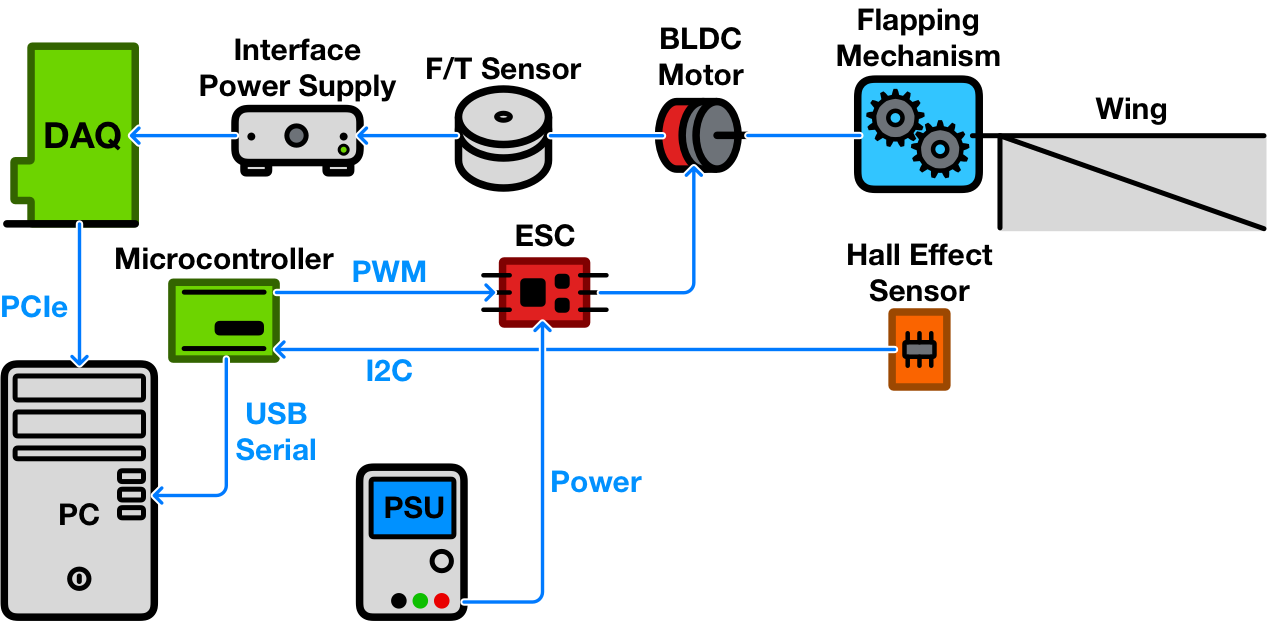}
    \caption{Overview of the experimental setup and data-acquisition pipeline for the propulsion unit.}
    \label{fig:ftpipeline}
\end{figure}

The ATI transducer is connected to its interface power supply box, which sends strain gauge voltage signals to the analogue-in ports of a PC-integrated NI PCIe-6320 DAQ card.
The propulsion unit’s BLDC motor is connected to a 0.4\,g Micro MX-5A ESC, which is powered by an external power supply unit (PSU) set at 3.7\,V to emulate a 1S battery. The ESC receives a throttle PWM signal from an Arduino Uno microcontroller, which is connected via USB to the data acquisition PC.
Additionally, the leading-edge-tracking sensor board is connected via I2C protocol to the microcontroller, which calculates the angular orientation of the wing (i.e. the flapping angle $\phi$) with a sensor polling rate of 317\,Hz and sends it via USB serial communication to the data acquisition PC. The full data-acquisition pipeline and the components of the experimental setup can be seen in Figure~\ref{fig:ftpipeline}.

On the data-acquisition PC, LabVIEW is used to calculate, synchronise, and export the test rig’s sensor data. A custom LabVIEW program transforms the strain gauge voltages read by the DAQ card via an ATI-supplied calibration matrix, resulting in a force and torque vector containing values for $F_x$, $F_y$, $F_z$, $T_x$, $T_y$, and $T_z$. These readouts can be tared and any offset corrected via a bias vector. The force and torque sensor data is acquired at 1000\,Hz and synchronised with the flapping angle data, with the result being recorded in a CSV file. Additionally, power usage is read off the PSU supplying the motor. In total, the test rig allows us to acquire force and torque data, the flapping angle, as well as current and voltage readings, from which electrical power usage, motor efficiency, and flapping frequency can be calculated.

\subsection{Pitch Limiter Testing Results} \label{pitchlimit}
As described in Section~\ref{flappingmechanism}, a passive wing-pitch mechanism was designed to enable a high degree of wing design freedom. With this system, a precise predefined wing-pitch angle $\theta_{max}$ can be set for the translation phases of the flapping cycle, while a revolute joint enables precise low-friction rotation of the whole wing during the stroke reversal.
To find an optimum in pitch angle for our design, an experimental approach was taken.
Several caps with varying endstop geometries were manufactured and can be screwed onto the wing-rotation joint, representing varying pitch limits. 
For this test series, five different pitch-angle limits were tested in 5º intervals, i.e., ±35º, ±40º, ±45º, ±50º, ±55º. For each version, three tests were performed and averaged with data acquired for 8\,s each at a throttle setting of 90\%.

\begin{figure}[htb]
    \centering
    \begin{subfigure}[t]{0.48\linewidth}
        \includegraphics[width=\linewidth]{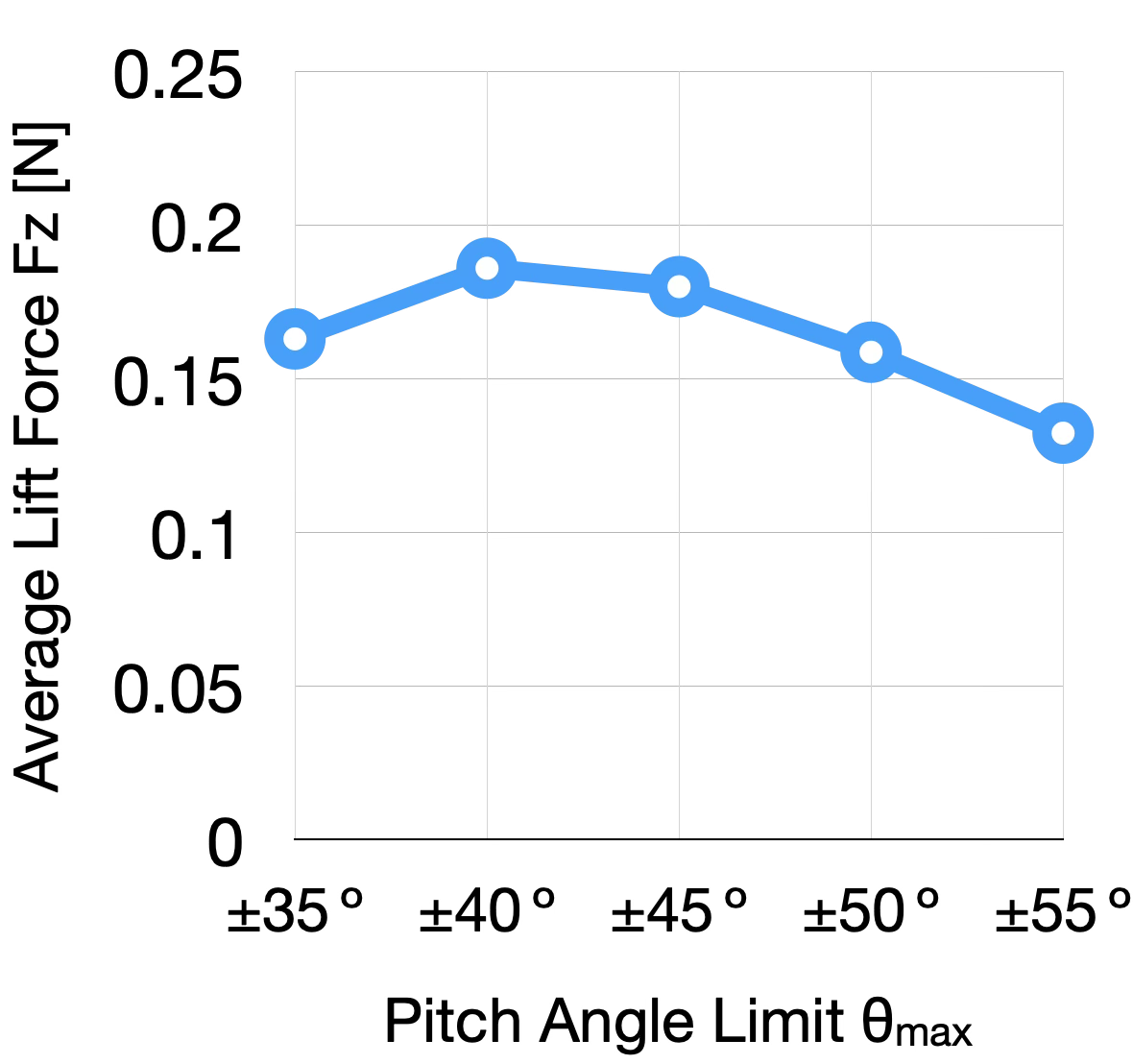}
        \caption{Average lift force $F_z$ generated for different pitch angle limits $\theta_{max}$.}
        \label{fig:pitchlimitfz}
    \end{subfigure}
    \hspace{0.01\linewidth}
    \begin{subfigure}[t]{0.48\linewidth}
        \includegraphics[width=\linewidth]{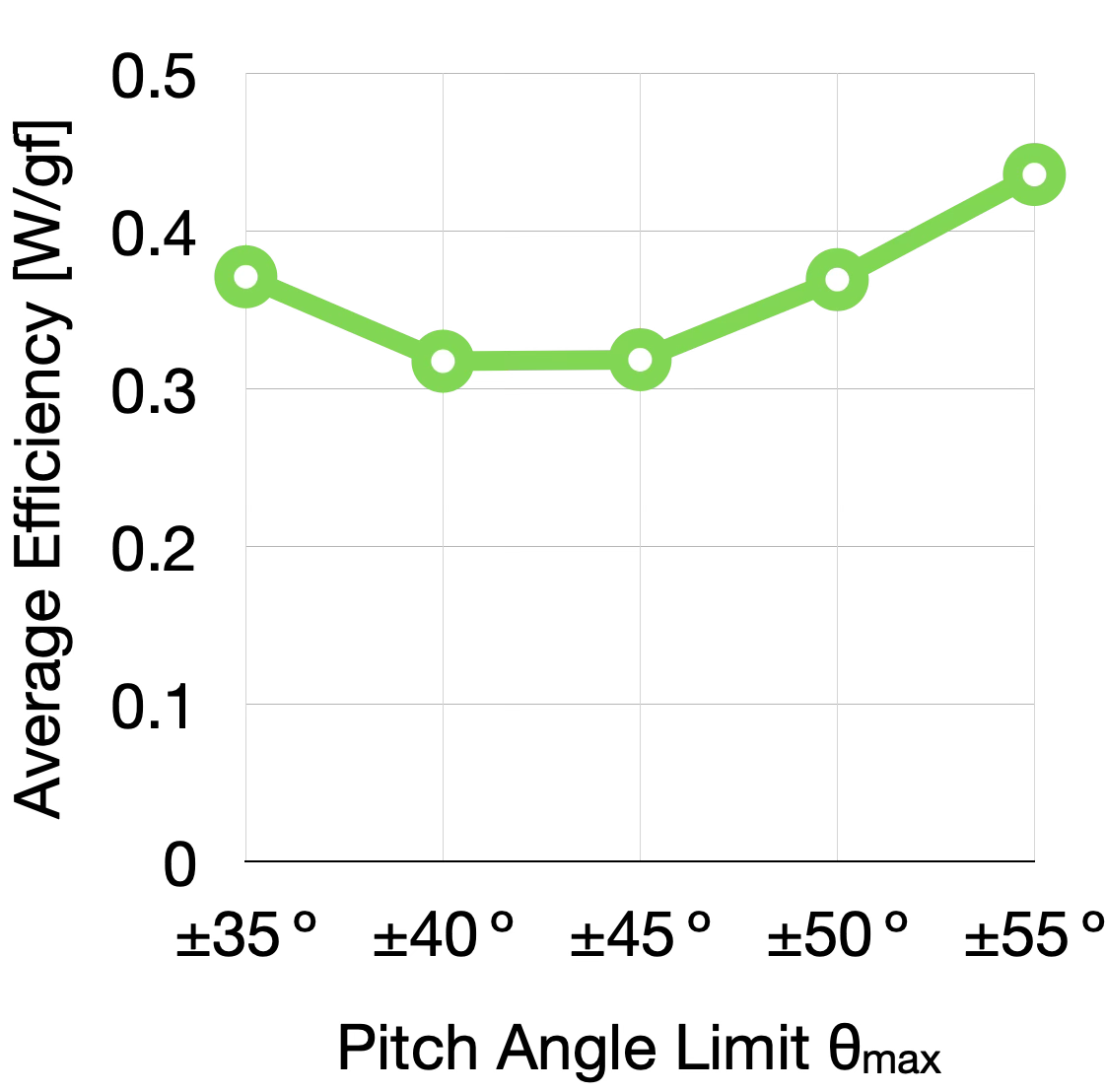}
        \caption{Average transmission efficiency of different pitch angle limits $\theta_{max}$.}
        \label{fig:pitchlimiteff}
    \end{subfigure}
    \caption{Test results for the five different pitch angle limiters of the passive wing-pitch mechanism.}
    \label{fig:pitchlimit}
\end{figure}

Of interest in finding the best design is especially the lift force generated, i.e. $F_z$, and the corresponding power usage to calculate the transmission efficiency for each pitch limiter design. Results of these tests can be seen in Figure~\ref{fig:pitchlimit}.

Regarding the average lift force generated, see Figure~\ref{fig:pitchlimitfz}, the ±40º pitch limiter achieved the highest forces at 0.186\,N, closely followed by the ±45º design at 0.18\,N, with all other tested variants performing significantly worse. A 5º deviation from the two best-performing pitch-angle limits resulted in an average loss in lift force of 12-13\%, while a 15º larger pitch angle than the optimum results in a drastic 29\% lift reduction.

For the efficiency, see Figure~\ref{fig:pitchlimiteff}, measured here as power required per gram-force lift; the ±40º and ±45º versions performed the same at 0.32\,W/gf, with all other versions again displaying worse efficiency. A 5º deviation from the two best-performing pitch-angle limits resulted in an efficiency loss of 15-17\%, while a 10º larger pitch angle compared to the most efficient configuration resulted in a loss of 37\%.

This testing successfully demonstrated the functionality of the passive pitch-limiter mechanism, and a clear optimum for our configuration was found in the ±40º pitch limiter, which was used in all subsequent tests. Additionally, it showed that even small deviations from the optimum wing pitch angle can result in significant performance losses.
In the literature, most FWMAVs utilise a degree of slack in the membrane with a fixed vertical root chord and don't specify the limits for their passive wing pitch. This degree of slack is represented by the angle between the root chord and the flapping axis; for example, the NUS-Roboticbird features a slack angle of only 5º~\cite{Nguyen2018}, while the Colibri implemented a slack angle of 16º~\cite{Roshanbin2017}.
The KU-Beetle utilised a 15º slack angle designed to achieve a ±50º pitch angle, as it proved more power efficient than a significantly shallower pitch angle according to a previous parametric study by them \cite{Phan2017,Phan2017a}.

Our precisely defined pitch-limiter design might enable future work to gather insights into general wing-pitch and wing-twist design guidelines.

\subsection{Throttle Range Testing Results}
\begin{figure}[htb]
    \centering
    \begin{subfigure}[t]{0.48\linewidth}
        \includegraphics[width=\linewidth]{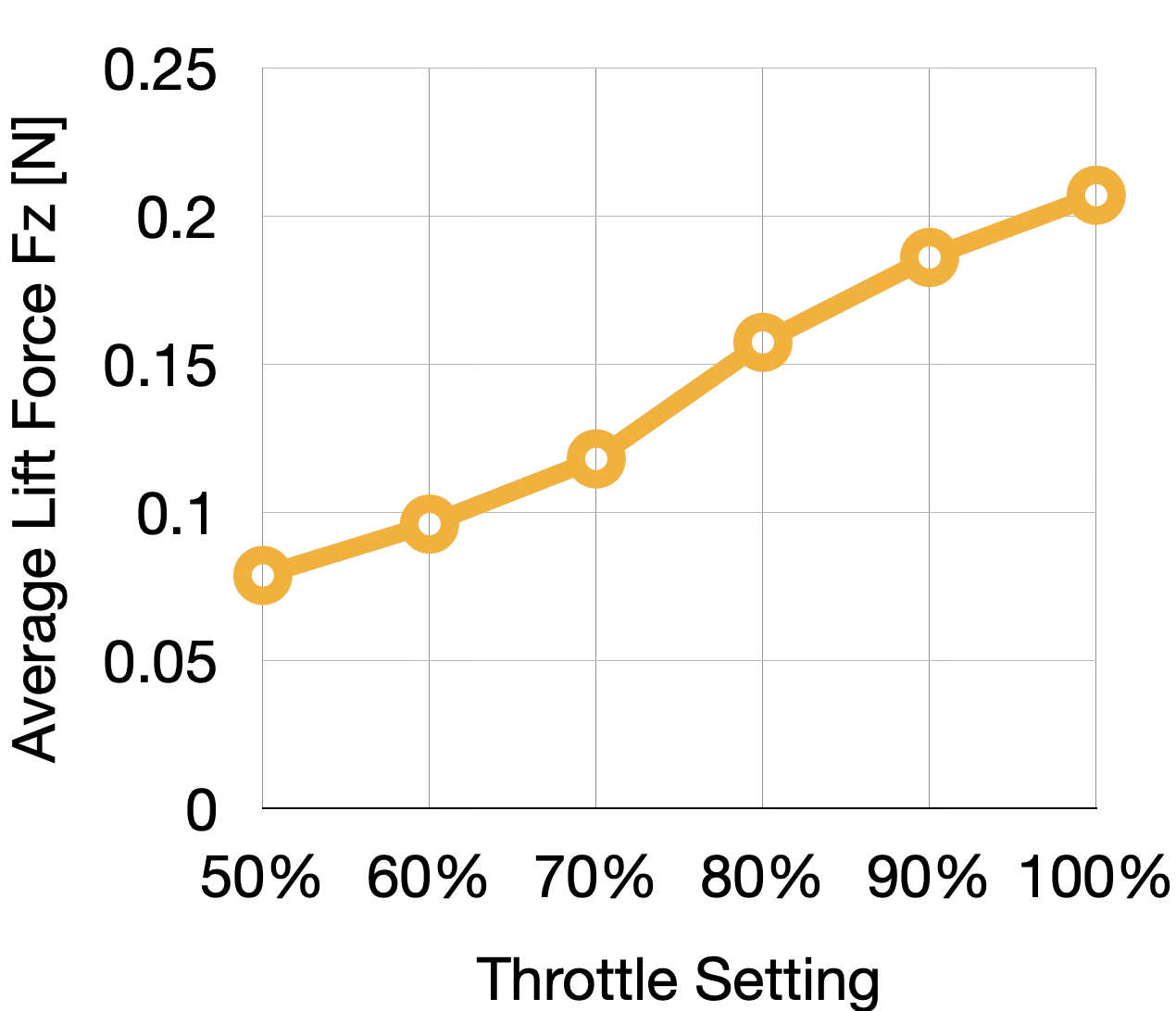}
        \caption{Average lift force $F_z$ generated over the throttle range showing a nearly linear response.}
        \label{fig:throttleset}
    \end{subfigure}
    \hspace{0.01\linewidth}
    \begin{subfigure}[t]{0.48\linewidth}
        \includegraphics[width=\linewidth]{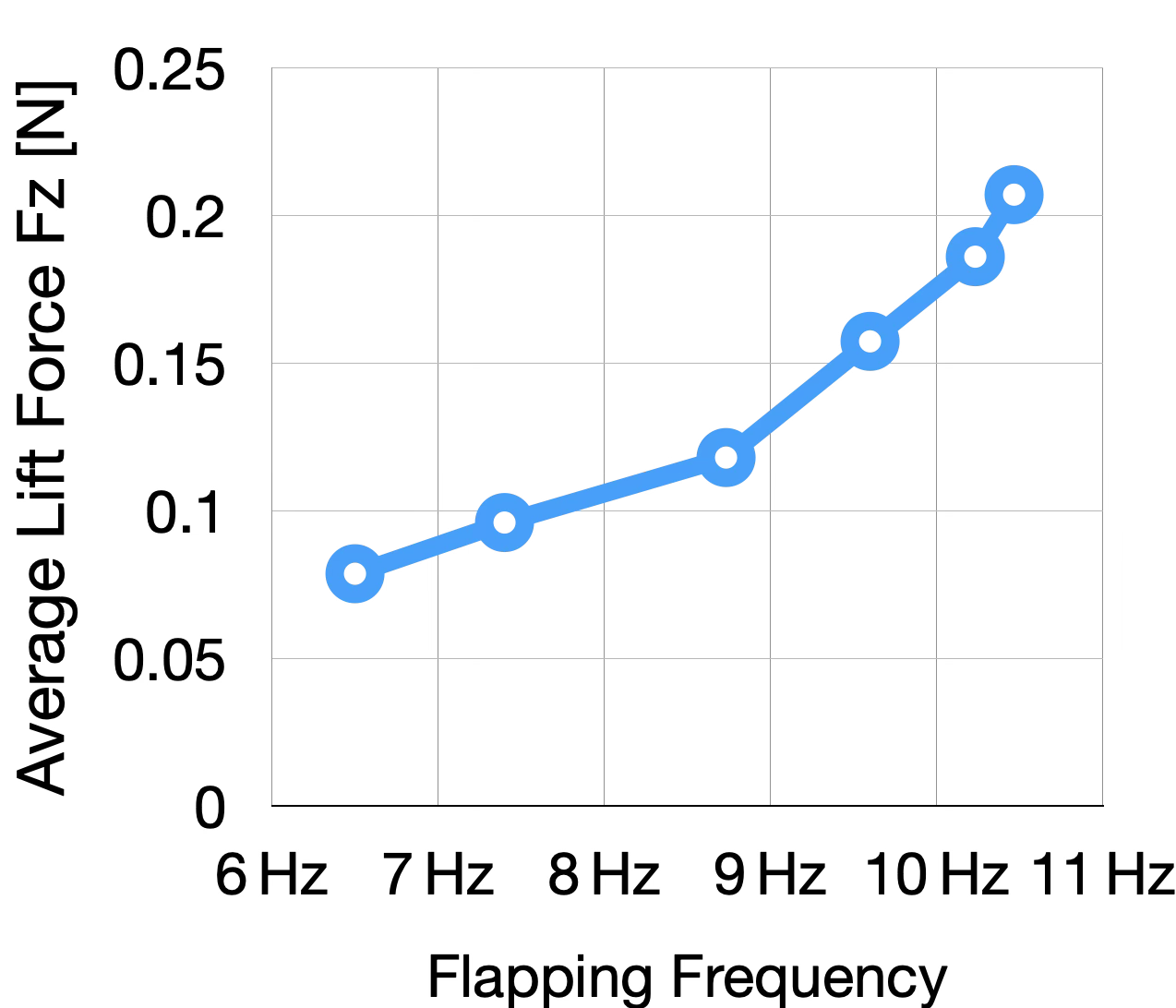}
        \caption{Average lift force $F_z$ generated in relation to the flapping frequency generated over the throttle range.}
        \label{fig:throttleflap}
    \end{subfigure}
    \caption{Results of the throttle response testing for a single-wing propulsion unit prototype.}
    \label{fig:throttletest}
\end{figure}

After finding the optimal pitch limit angle, the propulsion unit's throttle response curve was tested in the hover mode. For this, the throttle setting was changed from 50\% to 100\% in 10\% increments. At each setting, three test runs were performed for 8\,s each, and the results were averaged. In Figure~\ref{fig:throttletest}, the resulting throttle response curve is shown.

The average lift force generated, as seen in Figure~\ref{fig:throttleset}, shows a nearly linear response to throttle input, which will prove useful for control purposes.
The average maximum is reached at 0.207\,N for the 100\% throttle setting, which translates to 42.2\,gf generated for a two-winged configuration.

When the average lift force is plotted over the flapping frequency, see Figure~\ref{fig:throttleflap}, a characteristic quadratic graph is shown, similar to results reported by Preumont et al.~\cite{Preumont2021}. This also affirms our assumption made during transmission efficiency testing earlier, of flapping frequency as a useful proxy to gauge lift generation.
The maximum averaged flapping frequency reached at full throttle was 10.5\,Hz.

\begin{figure*}[tb]
    \centering
    \includegraphics[width=\linewidth]{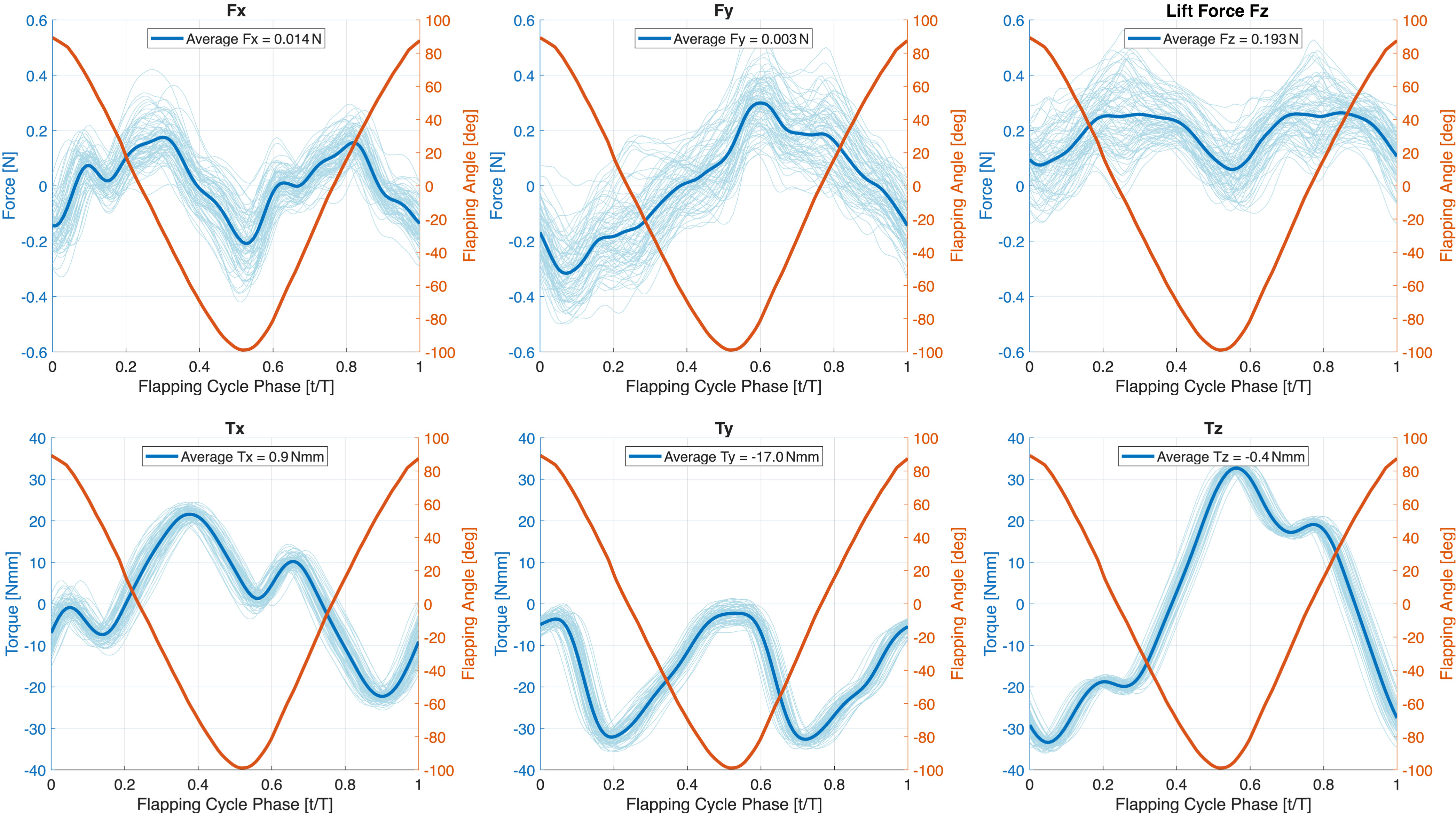}
    \caption{Flapping cycle averaged forces and torques measured in the sensor frame (see Figure~\ref{fig:ftsetup}) and overlaid with the average flapping angle from an 8\,s test with a 90\% throttle setting for a single-wing propulsion unit.
    The individual cycles of force and torque data contained in the single test data set are superimposed and displayed in the background in light blue}
    \label{fig:ftcycle}
\end{figure*}

To more closely analyse the forces and torques generated during the phases of the flapping cycle, the test series data were split at the points where the flapping angle reached a maximum, therefore signifying a completed flapping period of upstroke and downstroke. Each complete flapping cycle was then normalised by its period, i.e. $t/T$, and then superimposed on the previous cycles to calculate and display the average force and torque behaviour. To reduce noise, the force and torque data were low-pass filtered using a first-order Butterworth filter with a cut-off frequency of 20\,Hz, such that two flapping harmonics could be observed while attenuating higher-frequency noise. The unfiltered flapping-angle measurements were averaged for each data point and are displayed as a reference to the position during the flapping cycle. The results for an exemplary data series running for 8\,s with a 90\% throttle setting are shown in Figure~\ref{fig:ftcycle}.

\begin{figure}[htb]
    \centering
    \includegraphics[width=0.9\linewidth]{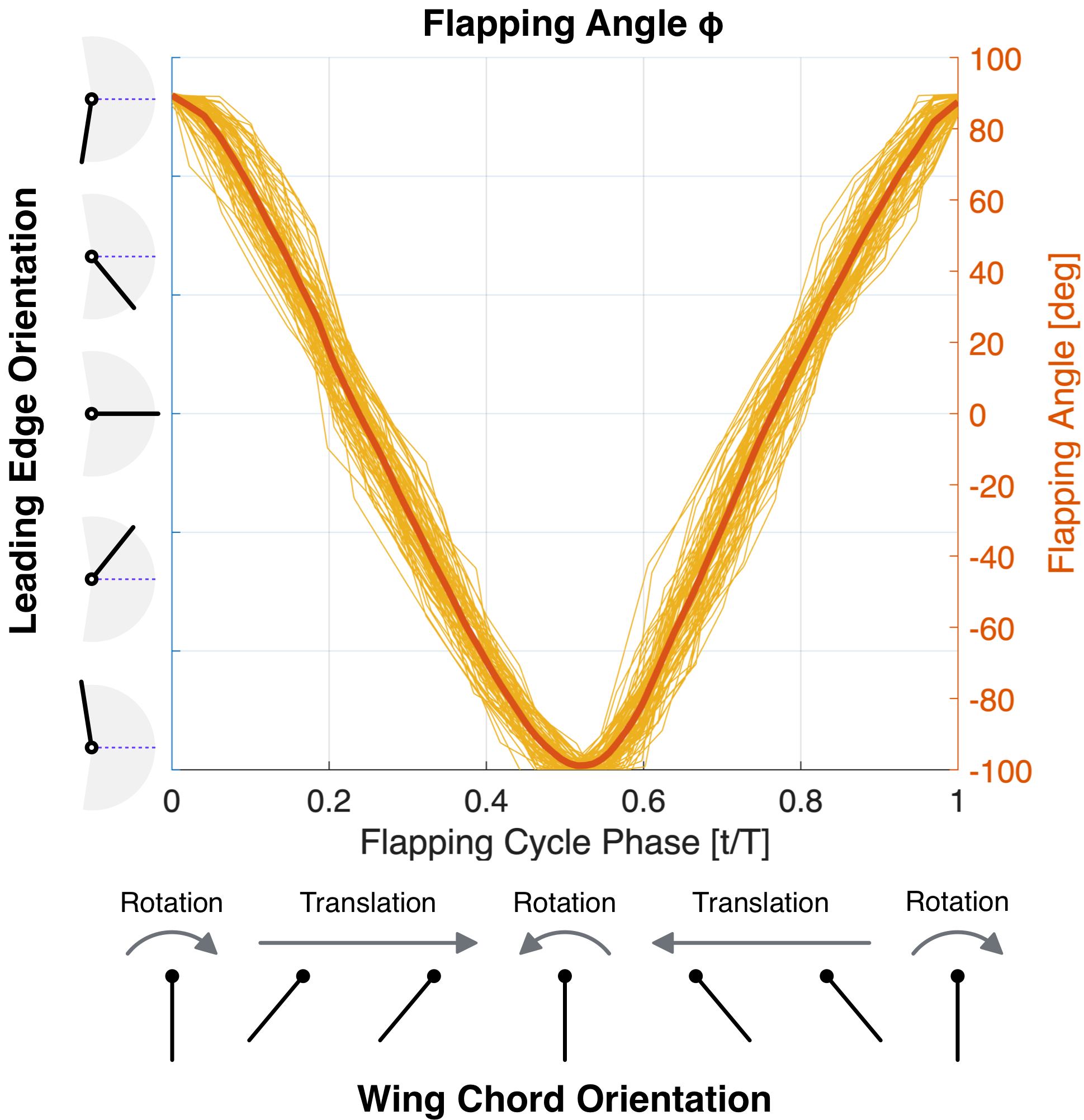}
    \caption{Flapping cycle normalised flapping angle $\phi$, showing the average track in orange, with individual unfiltered trajectories plotted in yellow. The exemplary data set from Figure~\ref{fig:ftcycle} is used from a single-wing propulsion unit running at 90\% throttle for 8\,s. Additionally shown to the left is a visual representation of the leading edge orientation within the stroke plane. Below, the corresponding wing-chord orientation and phases of the flapping cycle are displayed.}
    \label{fig:phicycle}
\end{figure}

Figure~\ref{fig:phicycle} shows the individual tracks of the flapping angle $\phi$ for the same data series and explains the cycle phases with the wing-chord and leading-edge orientation. Here, the average flapping angle shows a desired symmetry between upstroke and downstroke, with two nearly linear translation phases and two short rotational phases at each end of the stroke. The flapping cycle spans an average amplitude of 188º with a 10º zero-point offset due to manufacturing imprecision in the prototype, resulting in an averaged maximum of +89º and an averaged minimum of -99º flapping angle $\phi$.

The first panel of the top row in Figure~\ref{fig:ftcycle} shows the behaviour of the force $F_x$ measured in the $X_S$-direction of the sensor frame (see Figure~\ref{fig:ftsetup}). The symmetry of the cycle phases is evident, with a positive contribution during each translation phase and an opposing negative contribution during each rotation, resulting in a cycle-averaged force of only 0.014\,N. This small contribution is, however, fully neutralised in a two-winged configuration.
The second panel in the top row references the behaviour of the force $F_y$ in the direction of the $Y_S$-axis. It exhibits a negligible cycle-averaged contribution with a resulting average force of 0.003\,N.
The force of most interest is the lift force $F_z$ measured along the $Z_S$-axis, which is displayed in the third panel of the top row. Here, the translational and rotational phases can be clearly seen with symmetry between up and downstroke. During each translational phase, a nearly constant average lift of 0.26\,N is generated, which dips during the rotational phases to around 0.06\,N. This results in a cycle-averaged lift contribution of 0.193\,N, i.e. 19.7\,gf for a single wing. The final two-winged configuration is additionally expected to feature significant lift generation during the low-lift rotational phases due to utilisation of the clap-and-fling effect, enabled by the wide-angle transmission.
The bottom row of Figure~\ref{fig:ftcycle} features the torques generated around the sensor frame axes. In the first panel, the torque around the $X_S$-axis shows a small positive averaged contribution of 0.9\,Nmm, due to the asymmetry of the flapping angle in the current prototype.
The torque $T_y$ around the $Y_S$-axis features a significant contribution of -17\,Nmm, due to the significant lift force distribution acting on the leading edge of the wing. This torque contribution is neutralised by the opposing propulsion unit in the hovering flight condition.
Lastly, the torque $T_z$ around the $Z_S$-axis has an insignificant cycle-averaged contribution of -0.4\,Nmm due to the symmetry between upstroke and downstroke enabled by the flapping mechanism design.

\subsection{Tiltwing Testing and Control Moments Results}
\begin{figure}[htb]
    \centering
    \begin{subfigure}[t]{\linewidth}
        \centering
        \includegraphics[width=\linewidth]{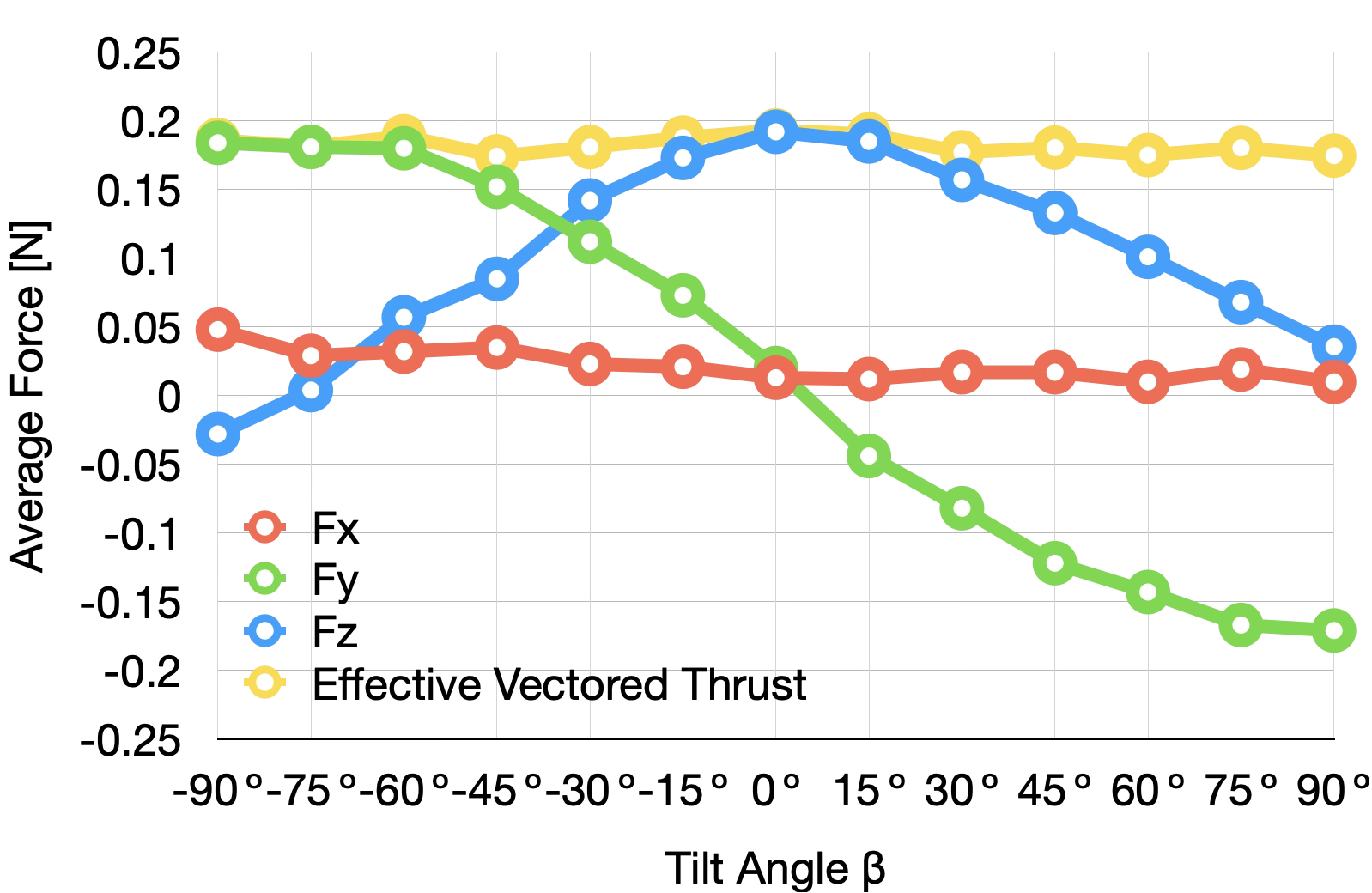}
        \caption{Experimental results for the average forces $F_x$,  $F_y$, $F_z$, and the average effective vectored thrust $F_T$ over the full range of the stroke-plane tilt angle $\beta$.}
        \label{fig:tiltforce}
    \end{subfigure}
    
    \vspace{1em}
    
    \begin{subfigure}[t]{\linewidth}
        \centering
        \includegraphics[width=\linewidth]{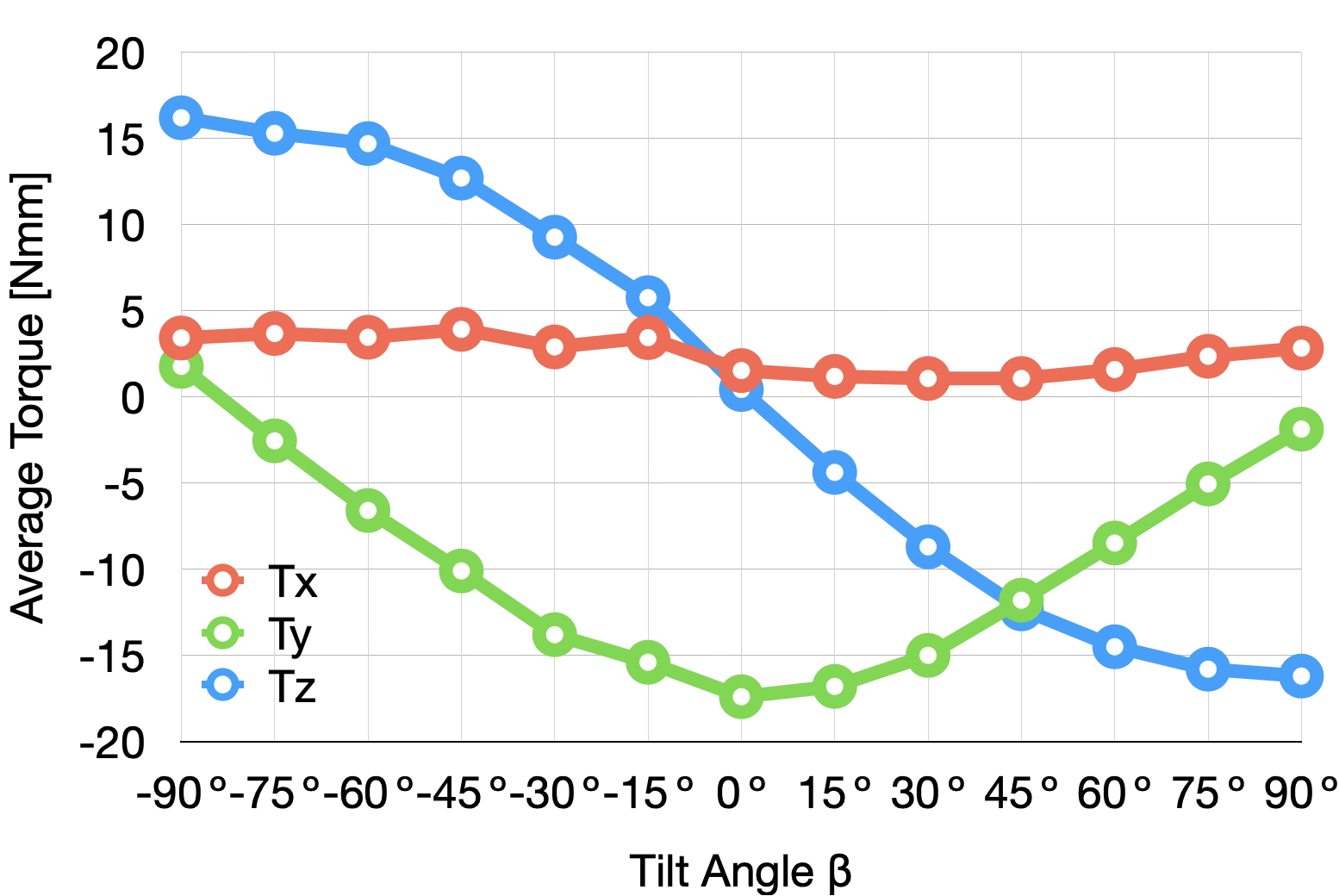}
        \caption{Experimental results for the average torques $T_x$, $T_y$, $T_z$ over the full range of the tilt angle $\beta$.}
        \label{fig:tilttorque}
    \end{subfigure}
    \caption{Loadcell test results for the tilted propulsion unit at a set thrust of 90\%.}
    \label{fig:tiltresults}
\end{figure}

In this test series, see Figure~\ref{fig:tiltresults}, the effects of the tiltwing capability are explored. At a static thrust setting of 90\%, the stroke-plane tilt angle $\beta$ is manually changed within the test rig to a desired degree by rotating the bridge holding the load cell and propulsion unit around its longitudinal axis. In Figure~\ref{fig:ftsetup}, the orientation of the tilting angle is shown, while the sensor frame with the effective measurement axes remains body-fixed for all measurements. The proposed configuration should be able to tilt its wings forward and backwards by 90º each; therefore, testing is conducted from -90º to +90º in 15º intervals, with 0º representing the horizontal stroke plane, i.e. the hover mode.

The resulting flapping cycle averaged forces in the sensor frame (see Figure~\ref{fig:ftsetup} for reference), i.e. $F_x$, $F_y$, $F_z$, are plotted over the tilt angle $\beta$ in Figure~\ref{fig:tiltforce}. Additionally, the effective vectored thrust $F_T$ is shown, demonstrating a consistent force output of 0.182\,N averaged across the flapping cycles, independent of the stroke-plane tilt angle $\beta$, suggesting reliable control output over the full tilting range.
This guarantees the conceptual advantage of our proposed design, which is capable of decoupling the wing orientation and thus its actuation system from the body attitude.
The lift force $F_z$ follows a slightly skewed cosine curve with $F_z \approx \cos(\beta) * T$ and a peak of 0.192\,N of average force produced. The perpendicular force $F_y$ follows a slightly skewed inverted sine curve with $F_y \approx -\sin(\beta) * T $ and a peak at 0.184\,N for a -90º angle as well as -0.171\,N for a +90º angle. This skew, especially visible in the $F_z$ curve, is due to the asymmetric flapping angle produced by the flapping-mechanism prototype, which exhibits a slightly off-centre zero point. This is due to assembly imprecision of the attachment between the rack and yoke, which will be alleviated in future models of the flapping mechanism with a defined connector geometry.

In Figure~\ref{fig:tilttorque}, the average generated torques in the sensor frame, i.e. $T_x$, $T_y$, $T_z$, are also plotted over the stroke-plane tilt angle $\beta$. They mirror the behaviour of the forces generated, as expected, with $T_y$ approximately following an inverted cosine curve and $T_z$ following a negative sine curve. $T_x$ shows the asymmetry of the current flapping mechanism prototype, as it produces 1.53\,Nmm of torque even at a 0º tilt angle. Significant torque is generated due to the thrust vectoring around the sensor's $Z_S$-axis, with peaks at ±16.2\,Nmm for ±90º tilt angle, contributing to the yawing moment. While the torque around the sensor's $Y_S$-axis peaks at -17.4\,Nmm for a 0º tilt, contributing to the rolling moment.

With this data, the behaviour of the configuration’s control architecture was calculated to give a first indication of the viability and capabilities of our approach. For this, realistic robot design parameters (see Table~\ref{tab:parameters}) were assumed, and the measured propulsion unit performance is mirrored to represent a two-winged design.

\begin{table}[htb]
\centering
\caption{FWMAV Design Parameters}
\label{tab:parameters}
\begin{tabular}{@{}lcc@{}}
\toprule
\multicolumn{1}{c}{Parameter}     & Symbol         & Value            \\ \midrule
Flapping frequency                & $f$            & 10.5\,Hz         \\
Flapping amplitude                & $2\phi_{max}$  & 190º             \\
stroke-plane tilt actuation angle & $\beta$        & ±90º             \\
Passive wing pitch angle          & $\theta_{max}$ & ±40º             \\
Target mass*                    & $m$            & $\approx$30\,g \\
Wing length                       & $l_{LE}$       & 130\,mm          \\
Chord length                      & $c$            & 60\,mm           \\
Centre of gravity Z-offset*       & $D_z$          & 30\,mm           \\
Centre of rotation Y-offset*      & $D_y$          & 20\,mm           \\
Motor mass                      & $m_{mot}$      & 3.1\,g           \\
Motor no load speed               & $n_0$          & 27,750\,rpm      \\
Transmission reduction ratio      & $i$            & 19$\times$       \\ \bottomrule
\multicolumn{3}{l}{\begin{tabular}[c]{@{}l@{}}*Realistic values assumed for the multibody simulation \\ and the calculation of expected control moments.\end{tabular}}
\end{tabular}
\end{table}

\begin{figure*}[tb]
    \centering
    \begin{subfigure}[t]{0.31\linewidth}
        \includegraphics[width=\linewidth]{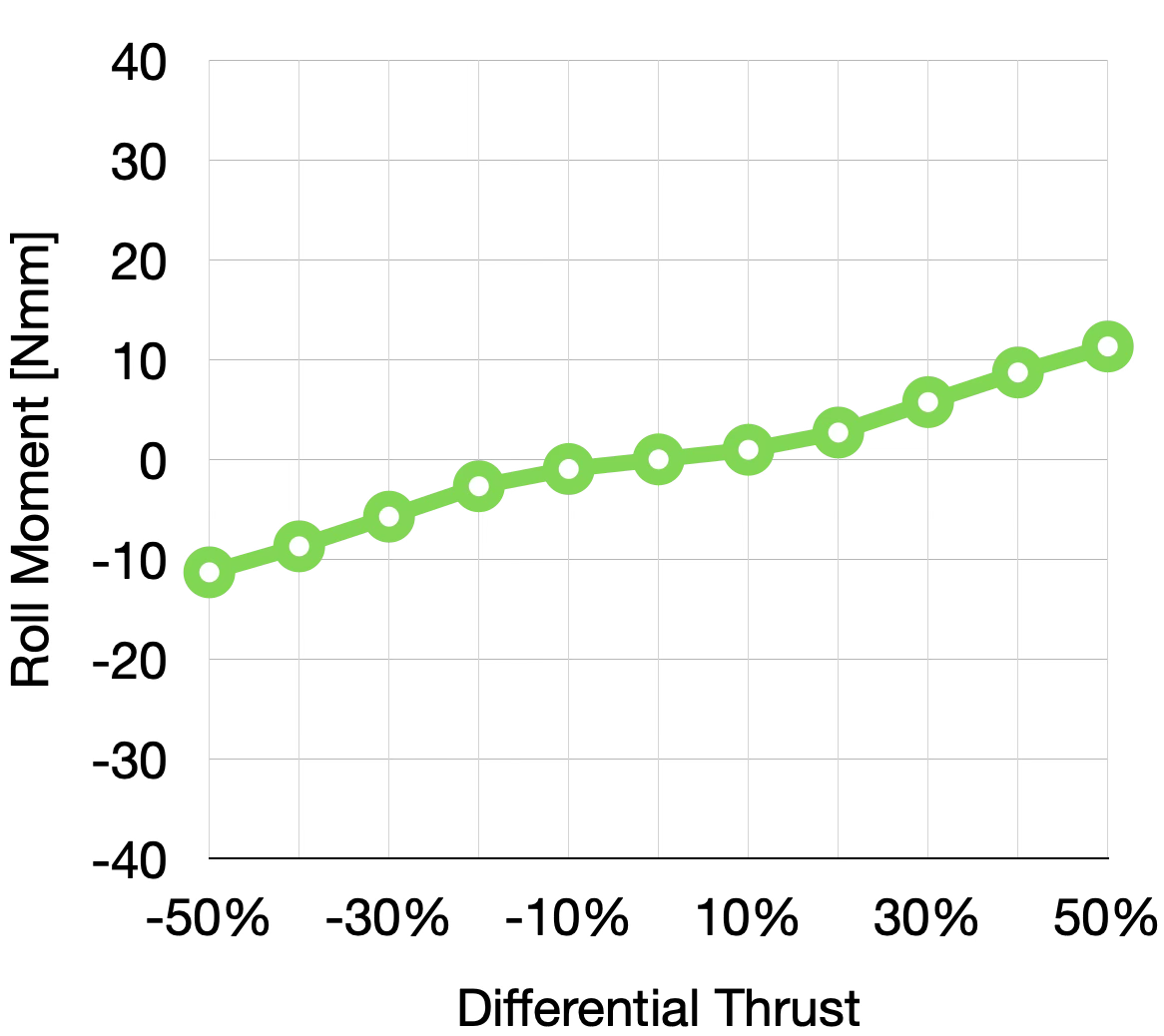}
        \caption{Calculated roll moment over differential thrust settings. The thrust of the right propulsion unit is subtracted from the thrust of the opposing left unit; thus, a positive thrust differential results in a positive roll moment.}
        \label{fig:rollefficacy}
    \end{subfigure}
    \hspace{0.01\linewidth}
    \begin{subfigure}[t]{0.31\linewidth}
        \includegraphics[width=\linewidth]{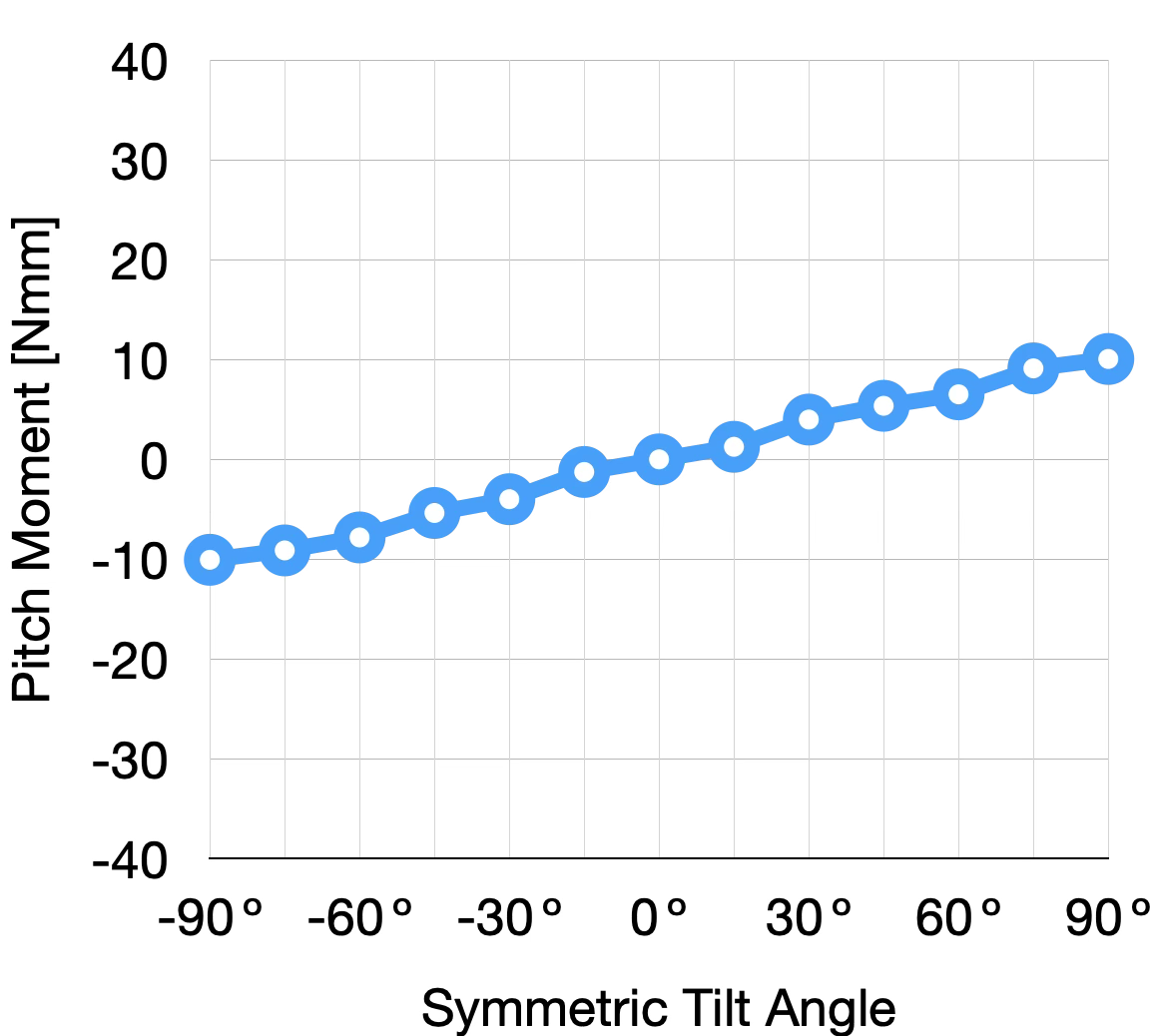}
        \caption{Calculated pitch moment over symmetric tilt angle setting. A positive pitch moment is generated for a positive tilt of both propulsion units around the right unit's $X_S$-axis.}
        \label{fig:pitchefficacy}
    \end{subfigure}
    \hspace{0.01\linewidth}
    \begin{subfigure}[t]{0.31\linewidth}
        \includegraphics[width=\linewidth]{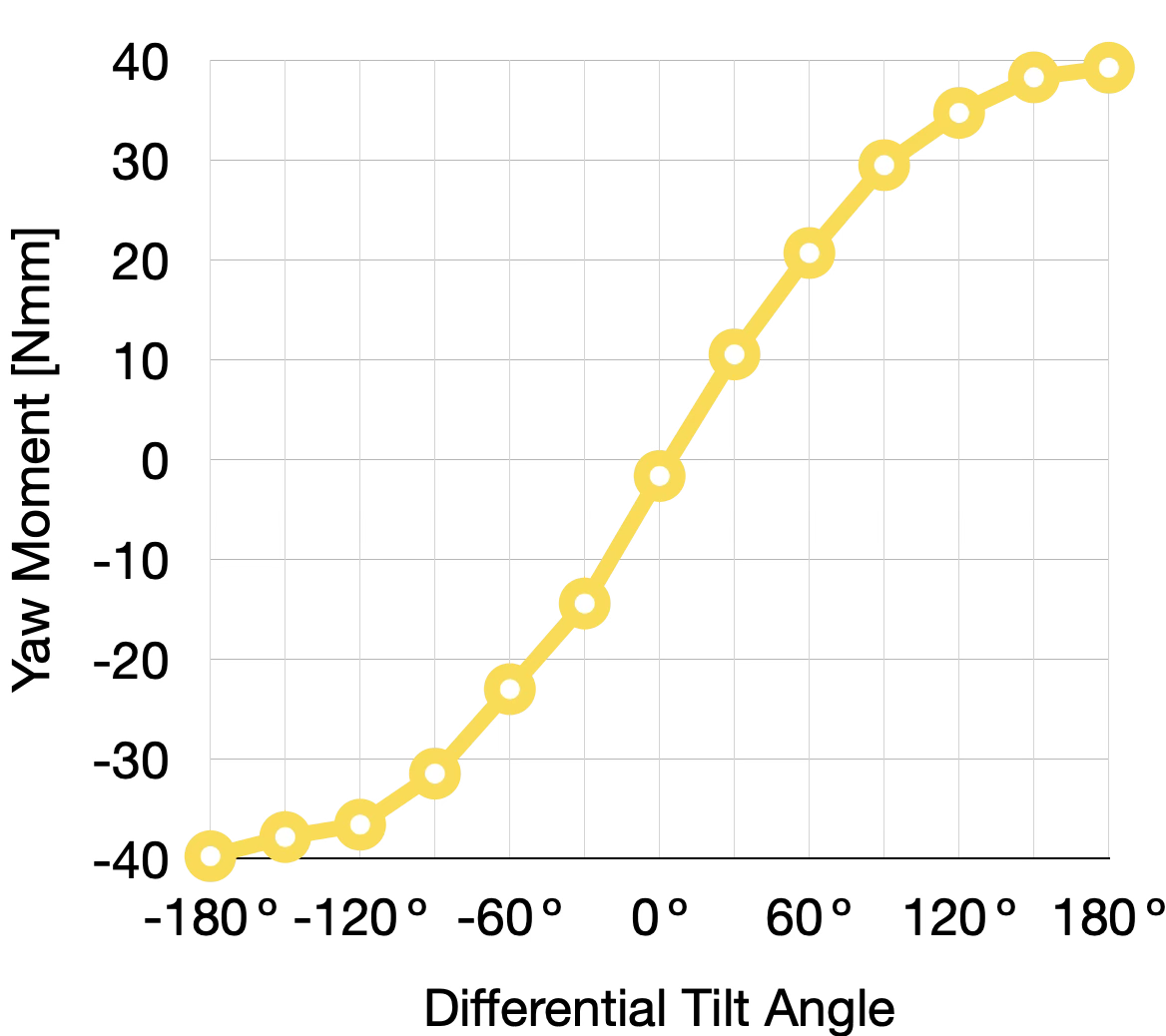}
        \caption{Calculated yaw moment over differential tilt angle. A positive yaw moment is generated for a positive tilt around the right unit's $X_S$-axis, with the left unit equally tilting in the opposite direction.}
        \label{fig:yawefficacy}
    \end{subfigure}
    \caption{Control moment efficacy for pitch, roll, and yaw calculated from tilt angle test data and throttle range test data, assuming a robot design with a defined centre-of-gravity offset $D_z = 30\,mm$ and centre-of-rotation offset $D_y = \pm20\,mm$.}
    \label{fig:controlmoments}
\end{figure*}

The propulsion units are assumed to be symmetrically mounted, as depicted in Figure~\ref{fig:fwcomponents}, with their respective centres-of-rotation offset $D_y$ by ±20\,mm along the robot body's $Y_B$-axis, to allow the physical space for each transmission. Additionally, they are situated above the centre of gravity by 30\,mm with the offset $D_z$ along the robot body's $Z_B$-axis to increase stability and enable effective control moment generation. With these two parameters, as well as the force and torque test data, control moments in the body-fixed frame for roll, pitch, and yaw are calculated and shown in Figure~\ref{fig:controlmoments}:
\begin{itemize}
    \item A desired rolling moment, see Figure~\ref{fig:rollefficacy}, is generated via differential throttle settings between the propulsion units. It can be calculated from the test data gathered during the throttle response testing by adding the rolling moment generated by the right propulsion unit to the rolling moment generated by its opposing left unit. 
    For this example, the thrust differential is calculated assuming one unit is always set to 100\% thrust as a baseline, and the thrust of the right unit is subtracted from the thrust of the left unit. Therefore, a positive roll moment is generated for a lower thrust of the right unit, see Figure~\ref{fig:hoverctrl}. 
    To calculate the rolling moment of the right unit, the lift force $F_z^R$ is multiplied by the lever arm $D_y$, i.e. the centre-of-rotation offset, and is then subtracted from the measured torque at the right propulsion unit $T_y^R$. To this, the span-wise-acting force $F_x^R$ multiplied by the lever arm $D_z$, i.e. the centre-of-gravity offset, is added.
    \begin{equation} 
        \begin{aligned}
            (T_x)_B =   & (-T_y^L-F_x^L*D_z+F_z^L*D_y)+\\
                        & +(T_y^R+F_x^R*D_z-F_z^R*D_y)
        \end{aligned}
    \end{equation}
 
    \item A desired pitching moment, see Figure~\ref{fig:pitchefficacy}, is generated by a symmetric tilting of both wings; therefore, tilting both thrust vectors in the same direction. It can be calculated by adding the individual pitching moments generated by each propulsion unit.
    For the right unit, the force $F_y^R$ is multiplied by the lever arm $D_z$, i.e. the centre-of-gravity offset, and subtracted from the torque $T_x^R$ at the right propulsion unit. A positive symmetric tilting around the right unit's $X_S$-axis generates a positive pitch moment, see Figure~\ref{fig:hoverctrl} and Figure~\ref{fig:ftsetup}.
    \begin{equation}
        (T_y)_B = (-T_x^L+F_y^L*D_z)+(T_x^R-F_y^R*D_z) 
    \end{equation}
    
    \item A desired yawing moment, see Figure~\ref{fig:yawefficacy}, is generated by a differential tilting of both wings’ stroke planes. It can be calculated by adding the individual yawing moments generated by each propulsion unit.
    For the right side, the force $F_y^R$ is multiplied by the lever arm $D_y$, i.e. the centre-of-rotation offset, and is subtracted from the inverse of the Torque $T_z^R$. A positive yaw moment is generated via a differential tilting angle with the right unit tilting around the positive $X_S$-axis, as seen in Figure~\ref{fig:hoverctrl} and Figure~\ref{fig:ftsetup}.
    \begin{equation} 
        (T_z)_B = (-T_z^L-F_y^L*D_y)+(-T_z^R-F_y^R*D_y)
    \end{equation}

\end{itemize}

With these calculations, the design parameters of the centre-of-gravity location and centre-of-rotation offset significantly influence the magnitude of the generated control moments, as they represent the respective lever arm for the produced forces. While the centre-of-rotation offset $D_y$ is mainly dependent on the transmission unit and frame size, the centre-of-gravity location $D_z$ will need to be carefully designed as it also affects the overall stability of the system in all flying modes.
A centre-of-gravity location below the wing attachment points will stabilise the system via the keel effect, with a limit of neutral stability at $D_z=0\,mm$ where the centre-of-gravity is located in the same plane as the generated forces.

The resulting roll moment of Figure~\ref{fig:rollefficacy} shows an effective, symmetric and nearly linear response, guaranteed by the underlying linear throttle response curve (see Figure~\ref{fig:throttleset}). For the measured range of ±50\% thrust differential, a rolling moment of ±11.3\,Nmm can be generated with the chosen design parameters.

The generated pitch moment, see Figure~\ref{fig:pitchefficacy}, displays an effective and symmetric linear response, due to the symmetric thrust vectoring, with a control output of ±10.1\,Nmm at the largest tilt settings. Even moderate tilt angles that guarantee sufficient lift generation, for example, a ±30º tilt, result in a significant pitch moment of ±4\,Nmm.

The yaw control output, see Figure~\ref{fig:yawefficacy}, is the most effective due to the configuration's thrust vectoring design and should allow for exceptional manoeuvrability. It exhibits a linear, nearly symmetric response curve that saturates for extremely large control inputs, reaching a maximum moment generated of +39.2\,Nmm and -39.8\,Nmm. A moderate tilt differential, for example, a delta in tilt angle of ±60º. results in a respective yawing moment of +23\,Nmm and -20.7\,Nmm. The slight asymmetry of the yaw control curve can be traced back to the asymmetry in the stroke plane of the prototype and will be remedied with improved manufacturing techniques in future work.

To compare these results to the other two FWMAV designs that use thrust vectoring (i.e.~\cite{Nguyen2018, Wu2026}), albeit with an X-Wing design, the moment arms were set to the same values used in these papers, allowing a fair comparison of the generated pitch and yaw moments.
Comparing the roll moments generated is harder, as roll control solely relies on the magnitude of the thrust differential between the left and right side, for which unavailable information on exact throttle settings would be required.
Table~\ref{tab:moments} gives an overview of the results of the following comparison.

The NUS-Roboticbird by Nguyen and Chan~\cite{Nguyen2018} featured a centre-of-gravity offset $D_z = 36\,mm$ and a centre-of-rotation offset $D_y = 10\,mm$ generating a maximum pitch moment of 1.7\,Nmm for a symmetric 15º stroke-plane tilt and a maximum yaw moment of 2.3\,Nmm for a differential 30º stroke-plane tilt. With the same moment arms, our design would generate a 2\,Nmm pitch moment and a 9.6\,Nmm yaw moment.

The X-Fly by Wu et al.~\cite{Wu2026} features an estimated centre-of-gravity offset $D_z \approx 25\,mm$ and an estimated centre-of-rotation offset $D_y \approx 55\,mm$. These values were estimated based on information provided in their paper and may not be highly accurate; however, they allow an approximation of the relative performance.
Their design generated a pitch moment of -6.1\,Nmm and +5.5\,Nmm for a symmetric ±15º stroke-plane tilt \cite{Wu2026}.
For a differential ±30º stroke-plane tilt, the robot generated a yaw moment of -7.4\,Nmm and +6.8\,Nmm \cite{Wu2026}. 
With their estimated moment arm parameters, our design would generate a ±4.2\,Nmm pitch moment, as well as a -13.7\,Nmm and +10.1\,Nmm yaw moment with these parameters.

\begin{table}[htb]
\centering
\caption{Control Moment Generation Comparison}
\label{tab:moments}
\begin{tabular}{@{}lcccc@{}}
\toprule
                                & \multicolumn{2}{c}{Nguyen and Chan~\cite{Nguyen2018}}                                              & \multicolumn{2}{c}{Our design}                                                  \\ \midrule
Pitch moment                    & \multicolumn{2}{c}{1.7\,Nmm}                                                     & \multicolumn{2}{c}{2\,Nmm}                                                      \\
Yaw moment                      & \multicolumn{2}{c}{2.3\,Nmm}                                                     & \multicolumn{2}{c}{9.6\,Nmm}                                                    \\ \midrule
\multicolumn{5}{l}{\begin{tabular}[c]{@{}l@{}}Pitch moment for a symmetric stroke-plane tilt of $\beta=+15^{\circ}$.\\ Yaw moment for a differential stroke-plane tilt of $\Delta\beta=+30^{\circ}$.\end{tabular}} \\
                                & \multicolumn{1}{l}{}                    & \multicolumn{1}{l}{}                   & \multicolumn{1}{l}{}                   & \multicolumn{1}{l}{}                   \\ \toprule
                                & \multicolumn{2}{c}{Wu et al.~\cite{Wu2026}}                                                    & \multicolumn{2}{c}{Our design}                                                                     \\ \midrule
Pitch moment                    & -6.1\,Nmm                               & 5.5\,Nmm                               & -4.2\,Nmm                              & 4.2\,Nmm                               \\
Yaw moment                      & -7.4\,Nmm                               & 6.8\,Nmm                               & -13.7\,Nmm                             & 10.1\,Nmm                              \\ \midrule
\multicolumn{5}{l}{\begin{tabular}[c]{@{}l@{}}Pitch moment for a symmetric stroke-plane tilt of $\beta=\pm 15^{\circ}$.\\ Yaw moment for a differential stroke-plane tilt of $\Delta\beta=\pm 30^{\circ}$.\end{tabular}}
\end{tabular}
\end{table}

Our configuration, with a similar motor size, produced equivalent results for the pitch moment compared to the already effective thrust-vectoring NUS-Roboticbird~\cite{Nguyen2018} and X-Fly~\cite{Wu2026}. However, our yaw actuation enabled a significantly larger control moment, suggesting a more effective implementation of the thrust vectoring design.
Furthermore, we tested a thrust-vectoring actuation range of ±90º, while the previously mentioned X-wing designs by Wu et al.~\cite{Wu2026} and Nguyen and Chan~\cite{Nguyen2018} feature a ±15º range and a ±20º range, respectively. This will enable the generation of larger control forces due to the availability of a wider actuation range, making more extreme manoeuvres possible compared to other FWMAV concepts.

Overall, we demonstrated that our design successfully implements thrust vectoring over a wide amplitude, with consistently generated forces and moments, which allows us to decouple wing actuation and body attitude. Our testing shows that this architecture represents a good fit for two-winged bioinspired aerial robots.
The most commonly implemented actuation and control architecture for tailless two-winged FWMAVs is wing-twist modulation (e.g.~\cite{Keennon2012,Phan2017,Roshanbin2017,Preumont2021}), which alters the lift produced by a wing in different sections of the cycle by modulating the root-chord orientation and therefore changing the slack in the wing membrane, leading to an altered maximum of the passive wing-pitch angle.
Our framework features two main advantages to this architecture:
We demonstrated effective and constant thrust production over a wide range of stroke-plane tilt angles, resulting in the generation of large predictable control forces. This allows the optimisation of the thrust production over the full flapping cycle (e.g. finding an optimal wing-pitch limit) compared to altering the lift production only locally to produce desired control moments.
Additionally, the reliance on control via root-chord modulation severely restricts the wing-design freedom, which our system circumvents, allowing further wing-design optimisations.

\subsection{Glide Mode Testing Results}
Following the quantification of the control forces generated during hovering flight, the functionality of the proposed glide-mode configuration was studied during initial tests.
In this flight mode, the robot acts as a glider with active wing-incidence-angle and anhedral/dihedral-angle control. For this, the motor switches from its normal continuous flapping operation to a closed-loop servo operation to allow the active setting of a desired anhedral/dihedral angle or a neutral gliding position of the wing. The motor’s ESC has been configured to allow bidirectional output depending on the PWM signal sent, leading to the motor being able to reverse without needing to complete a full stroke.
With the angular-position feedback of the motor's Hall-effect sensor, a simple P-controller was designed and tuned. After preliminary experimentation, a ±10º deadzone was implemented, achieving good accuracy while preventing oscillations. Under a wind load, several dihedral- and anhedral-angle commands were given, resulting in quick position changes and a general accuracy of under 5º of deviation during repeated testing.
Some performance challenges were noted when commanding smaller angle changes, which can easily lead to overshoots and corrections, as the current transmission design features large static friction, which requires large startup speeds. These are also due to the lower reduction rate when compared to a conventional servo motor. With future improvements in transmission tolerances and manufacturing, these challenges will be addressed for better tracking accuracy.

\begin{figure}[htb]
    \centering
    \includegraphics[width=0.7\linewidth]{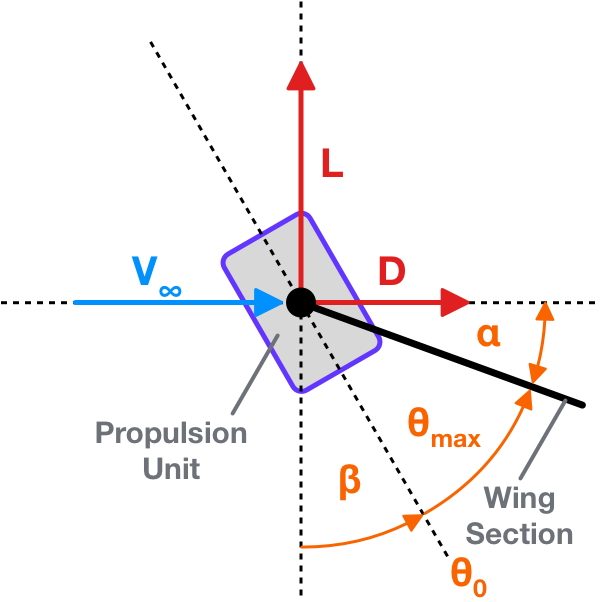}
    \caption{Chord-side view of the wing in the glide mode. The propulsion unit is tilted into the airstream $V$ with the angle $\beta$, while the wing is pitched under this load to its maximum pitch angle $\theta_{max}$, resulting in an effective angle of attack $\alpha$. The tilting capability of the proposed design, therefore, actively controls the angle of attack during glide for each wing individually and thus the lift and drag generated.}
    \label{fig:glidechord}
\end{figure}

The other active wing-adjustment system is represented by the active wing-incidence-angle control, i.e. the wing-tilt control, which effectively can adjust the angle of attack for each wing independently and is proposed as the main control input for the glide mode.
In these initial tests, we present an unoptimised gliding design that only aims to demonstrate the general functionality of this mode, serving as a proof-of-concept, which leaves major improvements and the implementation of a controller for flight tests for future work.
To test the effectiveness of the wing-tilt control in the gliding mode, a large fan was introduced into the test setup to provide an incoming airstream.
A viable configuration in this mode is achieved when the stroke plane is tilted into the wing, and the tilt angle $\beta$ is adjusted with consideration of the passive pitch-angle limit $\theta_{max}$, which is set to ±40º with the current design. 
For an incoming flow parallel to the body’s $X_B$-axis, the theoretical angle of attack can then be calculated as $\alpha = 90 - \theta_{max} - \beta$ for a dihedral angle of 0º, as shown in Figure~\ref{fig:glidechord}.

\begin{figure}[htb]
    \centering
    \includegraphics[width=0.7\linewidth]{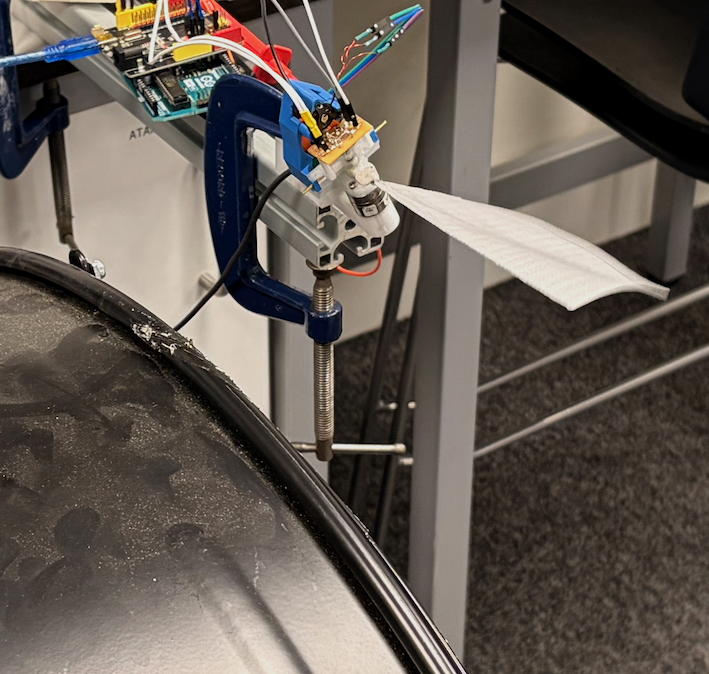}
    \caption{Tilted propulsion unit installed on a load cell during preliminary glide-performance testing under wind load from a large fan.}
    \label{fig:fan}
\end{figure}

In this initial test series, see Table~\ref{tab:glidetest}, the wing-tilt angle $\beta$ is manually set to a value between 25º and 50º in 5º intervals, resulting in a theoretical angle of attack $\alpha$ of 25º to 0º. The motor controller was set to a 0º dihedral configuration for all tests, resulting in an average measured dihedral angle of 1º.
The fan used in the test setup, see Figure~\ref{fig:fan}, produced an average airspeed of 6\,m/s at the wing with a variance of roughly ±0.2\,m/s, which was verified using a hot-wire anemometer.

Forces sensed by the force/torque sensor were transformed into lift and drag force components to calculate the resultant force and the glide angle $\gamma$ for each tilt-angle setting. From this, the supported robot mass was calculated for an ideal glide path at the set speed.

\begin{table}[htb]
\centering
\caption{Glide Mode Testing Results}
\label{tab:glidetest}
\begin{tabular}{@{}ccccccc@{}}
\toprule
\begin{tabular}[c]{@{}c@{}}Tilt \\ Angle \\ $\beta$\end{tabular}             & \begin{tabular}[c]{@{}c@{}}AoA \\ $\alpha$\end{tabular}             & \begin{tabular}[c]{@{}c@{}}Average \\ Lift\end{tabular}            & \begin{tabular}[c]{@{}c@{}}Resultant \\ Force\end{tabular}            & \begin{tabular}[c]{@{}c@{}}Supported \\ Robot \\ Mass\end{tabular}            & \begin{tabular}[c]{@{}c@{}}Glide \\ Angle $\gamma$\end{tabular}            & \begin{tabular}[c]{@{}c@{}}Glide \\ Ratio\end{tabular}            \\ \midrule
25º                                                                                  & 25º                                                                 & 0.174\,N                                                           & 0.223\,N                                                              & 45.4\,g                                                                         & 38.6º                                                                      & 1.3                                                               \\
30º                                                                                  & 20º                                                                 & 0.157\,N                                                           & 0.204\,N                                                              & 41.7\,g                                                                         & 39.8º                                                                      & 1.2                                                               \\
35º                                                                                  & 15º                                                                 & 0.163\,N                                                           & 0.197\,N                                                              & 40.2\,g                                                                         & 34.3º                                                                      & 1.5                                                               \\
40º                                                                                  & 10º                                                                 & 0.167\,N                                                           & 0.189\,N                                                              & 38.6\,g                                                                         & 27.9º                                                                      & 1.9                                                               \\
45º                                                                                  & 5º                                                                  & 0.152\,N                                                           & 0.169\,N                                                              & 34.5\,g                                                                         & 26º                                                                        & 2.1                                                               \\
50º                                                                                  & 0º                                                                  & 0.138\,N                                                           & 0.152\,N                                                              & 31.1\,g                                                                         & 25.1º                                                                      & 2.1                                                               \\ \midrule
\multicolumn{7}{l}{\begin{tabular}[c]{@{}l@{}}Test results for the glide mode of a single wing with an airspeed of 6\,m/s \\ and an average measured dihedral angle of 1º with a standard deviation \\ of 0.2º between tilt settings. With the passive pitch angle $\theta_{max}$ at 40º, \\ different tilt angle settings $\beta$ result in a defined angle of attack $\alpha$. \\ The average lift and resultant force were measured for a single wing, \\ with the supported robot mass being calculated for two wings.\end{tabular}}
\end{tabular}
\end{table}

The currently implemented unoptimised wing design displays a poor glide performance with a glide ratio of around 2, requiring future wing-design improvements to achieve a more efficient gliding flight.
However, testing showed that the current design was able to support the initially targeted robot mass of 30\,g. Furthermore, it showed that small changes in the wing-tilt angle resulted in significant changes to the lift and drag forces produced, suggesting reliable control surfaces during gliding.

For a more detailed investigation of this gliding mode, future work will focus on wing design and actuation-accuracy improvements. Additionally, wind tunnel testing will be required to further quantify the wing behaviour and the control moments generated during glide for varying wing-tilt angles and dihedral/anhedral angles.

\section{Conclusion}\label{conclusion}
In this paper, we presented a novel multimodal tiltwing framework for bioinspired aerial robots that relies on effective thrust-vectoring control. It aims to allow flapping-wing MAVs to perform hovering flight, high-speed directional flight, and gliding flight, with enhanced manoeuvrability and efficiency. By learning from nature's most proficient flyers, this extension of the available flight envelope will substantially increase the capabilities and range of application for aerial robots.
To meet the specific requirements of this approach, a hybrid Scotch-yoke-based flapping mechanism was developed with the help of a multibody-physics simulation, and extensively tested in hardware iterations.
This mechanism features an easily configurable design that achieves arbitrarily large flapping amplitudes, often a challenge in existing designs, enabling the utilisation of the lift-enhancing clap-and-fling effect for two-winged FWMAVs.
Furthermore, it produces a symmetric motion profile, highly desired for stable two-winged operation.
Additionally, an accurate leading-edge tracking sensor was implemented, enabling the use of two independent propulsion units and the switch to a gliding mode with active wing-pitch and dihedral-angle control.

The forces and torques generated by a single propulsion unit were characterised, and an optimal configuration of the transmission system and passive wing-rotation mechanism was found.
The single-wing prototype generated an average of 21.1\,gf of lift at full throttle with a 3.1\,g 1S BLDC motor, using an unoptimised and simplified wing design.
With the testing of our ±90º stroke-plane tilting actuation range of the framework, we demonstrated the effectiveness of our thrust-vectoring approach for two-winged FWMAVs.
Resulting control moments were calculated, showing a linear and symmetric response curve with great moment generation for its design size, which should enable advanced manoeuvrability, compared to other concepts.
Further architectural advantages of our framework comprise an uncoupling of body attitude and wing actuation, as well as providing extensive freedom in wing design, allowing future optimisation for thrust production over the full flapping cycle.

Lastly, the functionality of the glide mode of the system was demonstrated in a preliminary test setup, showing the active anhedral/dihedral and pitch-angle control. Switching to a glide mode is facilitated through our use of an omnidirectional BLDC motor in combination with the contactless leading-edge tracking sensor. Here, the dual-wing architecture also represents an advantage as it simplifies wing design for optimised glide performance. 

In future work, the prototype of the propulsion unit will be overhauled for precise and repeatable manufacturing, reducing friction and backlash in the transmission design. With two weight-optimised units and two tilting servos, a flight-capable robot will be assembled, and the proposed control architecture fully implemented and tested.
Challenges will be in effective synchronisation of both wings to suppress unwanted oscillations and achieve stable flight.
Additional optimisation will be required to exploit the wing design freedom enabled by our framework. Here, thrust generation over the full flapping phase and efficiency of the wing with consideration of the different flight modes will need to be addressed.

\bibliographystyle{ieeetr}
\bibliography{References}

@article{Armanini2019,
  title = {Modelling {{Wing Wake}} and {{Tail Aerodynamics}} of a {{Flapping-Wing Micro Aerial Vehicle}}},
  author = {Armanini, Sf and Caetano, Jv and De Visser, Cc and Pavel, Md and De Croon, Gche and Mulder, M},
  year = 2019,
  month = jan,
  journal = {International Journal of Micro Air Vehicles},
  volume = {11},
  pages = {1756829319833674},
  issn = {1756-8293, 1756-8307},
  doi = {10.1177/1756829319833674},
  urldate = {2026-05-18},
  langid = {english}
}

@article{Brodsky1991,
  title = {Vortex {{Formation}} in the {{Tethered Flight}} of the {{Peacock Butterfly Inachis IO L}}. ({{Lepidoptera}}, {{Nymphalidae}}) and {{Some Aspects}} of {{Insect Flight Evolution}}},
  author = {Brodsky, A. K.},
  year = 1991,
  month = nov,
  journal = {Journal of Experimental Biology},
  volume = {161},
  number = {1},
  pages = {77--95},
  issn = {0022-0949, 1477-9145},
  doi = {10.1242/jeb.161.1.77},
  urldate = {2026-04-21},
  copyright = {http://www.biologists.com/user-licence-1-1/},
  langid = {english}
}

@article{Broers2025,
  title = {Repeatable {{Energy-Efficient Perching}} for {{Flapping-Wing Robots Using Soft-Grippers}}},
  author = {Broers, Krispin C V and Armanini, Sophie F},
  year = 2025,
  month = nov,
  journal = {Bioinspiration \& Biomimetics},
  volume = {20},
  number = {6},
  pages = {066017},
  issn = {1748-3182, 1748-3190},
  doi = {10.1088/1748-3190/ae18a8},
  urldate = {2026-04-10},
  langid = {english}
}

@article{Chin2016,
  title = {Flapping {{Wing Aerodynamics}}: {{From Insects}} to {{Vertebrates}}},
  shorttitle = {Flapping {{Wing Aerodynamics}}},
  author = {Chin, Diana D. and Lentink, David},
  year = 2016,
  month = apr,
  journal = {Journal of Experimental Biology},
  volume = {219},
  number = {7},
  pages = {920--932},
  issn = {1477-9145, 0022-0949},
  doi = {10.1242/jeb.042317},
  urldate = {2024-11-19},
  copyright = {http://www.biologists.com/user-licence-1-1},
  langid = {english}
}

@article{Coleman2017,
  title = {Development of a {{Robotic Hummingbird Capable}} of {{Controlled Hover}}},
  author = {Coleman, David and Benedict, Moble and Hirishikeshaven, Vikram and Chopra, Inderjit},
  year = 2017,
  month = jul,
  journal = {Journal of the American Helicopter Society},
  volume = {62},
  number = {3},
  pages = {1--9},
  issn = {2161-6027},
  doi = {10.4050/JAHS.62.032003},
  urldate = {2025-01-15},
  langid = {english}
}

@article{DeCroon2012,
  title = {Design, {{Aerodynamics}} and {{Autonomy}} of the {{Delfly}}},
  author = {De Croon, G C H E and Groen, M A and De Wagter, C and Remes, B and Ruijsink, R and Van Oudheusden, B W},
  year = 2012,
  month = jun,
  journal = {Bioinspiration \& Biomimetics},
  volume = {7},
  number = {2},
  pages = {025003},
  issn = {1748-3182, 1748-3190},
  doi = {10.1088/1748-3182/7/2/025003},
  urldate = {2025-08-08},
  langid = {english}
}

@article{Deng2020,
  title = {Design {{Optimization}} and {{Experimental Study}} of a {{Novel Mechanism}} for a {{Hover-Able Bionic Flapping-Wing Micro Air Vehicle}}},
  author = {Deng, Huichao and Xiao, Shengjie and Huang, Binxiao and Yang, Lili and Xiang, Xinyi and Ding, Xilun},
  year = 2020,
  month = mar,
  journal = {Bioinspiration \& Biomimetics},
  volume = {16},
  number = {2},
  pages = {026005},
  issn = {1748-3182, 1748-3190},
  doi = {10.1088/1748-3190/abc292},
  urldate = {2026-05-20}
}

@article{Dickinson1999,
  title = {Wing {{Rotation}} and the {{Aerodynamic Basis}} of {{Insect Flight}}},
  author = {Dickinson, Michael H. and Lehmann, Fritz-Olaf and Sane, Sanjay P.},
  year = 1999,
  month = jun,
  journal = {Science},
  volume = {284},
  number = {5422},
  pages = {1954--1960},
  issn = {0036-8075, 1095-9203},
  doi = {10.1126/science.284.5422.1954},
  urldate = {2025-08-12},
  langid = {english}
}

@article{Dickson2008,
  title = {Integrative {{Model}} of {{Drosophila Flight}}},
  author = {Dickson, William B. and Straw, Andrew D. and Dickinson, Michael H.},
  year = 2008,
  month = sep,
  journal = {AIAA Journal},
  volume = {46},
  number = {9},
  pages = {2150--2164},
  issn = {0001-1452, 1533-385X},
  doi = {10.2514/1.29862},
  urldate = {2025-08-06},
  langid = {english}
}

@misc{Festo2024,
  title = {{{BionicBee}}: {{Autonomous Flying}} in a {{Swarm}}},
  author = {Festo},
  year = 2024
}

@inproceedings{Grasmeyer2001,
  title = {Development of the {{Black Widow Micro Air Vehicle}}},
  booktitle = {39th {{Aerospace Sciences Meeting}} and {{Exhibit}}},
  author = {Grasmeyer, Joel and Keennon, Matthew},
  year = 2001,
  month = jan,
  publisher = {{American Institute of Aeronautics and Astronautics}},
  address = {Reno,NV,U.S.A.},
  doi = {10.2514/6.2001-127},
  urldate = {2026-04-10},
  langid = {english}
}

@book{Gundlach2012,
  title = {Designing {{Unmanned Aircraft Systems}}: {{A Comprehensive Approach}}},
  shorttitle = {Designing {{Unmanned Aircraft Systems}}},
  author = {Gundlach, Jay},
  year = 2012,
  month = jan,
  publisher = {{American Institute of Aeronautics and Astronautics}},
  address = {Reston ,VA},
  doi = {10.2514/4.868443},
  urldate = {2025-08-08},
  isbn = {978-1-60086-843-6 978-1-60086-844-3},
  langid = {english}
}

@article{Guo2024,
  title = {Development of a {{Bio-Inspired Tailless Fwmav}} with {{High-Frequency Flapping Wings Trajectory Tracking Control}}},
  author = {Guo, Qingcheng and Wu, Chaofeng and Zhang, Yichen and Cui, Feng and Liu, Wu and Wu, Xiaosheng and Lu, Junguo},
  year = 2024,
  month = sep,
  journal = {Journal of Bionic Engineering},
  volume = {21},
  number = {5},
  pages = {2145--2166},
  issn = {1672-6529, 2543-2141},
  doi = {10.1007/s42235-024-00554-y},
  urldate = {2025-02-19},
  langid = {english}
}

@article{Karasek2018,
  title = {A {{Tailless Aerial Robotic Flapper Reveals That Flies Use Torque Coupling}} in {{Rapid Banked Turns}}},
  author = {Kar{\'a}sek, Mat{\v e}j and Muijres, Florian T. and De Wagter, Christophe and Remes, Bart D. W. and De Croon, Guido C. H. E.},
  year = 2018,
  month = sep,
  journal = {Science},
  volume = {361},
  number = {6407},
  pages = {1089--1094},
  issn = {0036-8075, 1095-9203},
  doi = {10.1126/science.aat0350},
  urldate = {2025-01-17},
  langid = {english}
}

@inproceedings{Keennon2012,
  title = {Development of the {{Nano Hummingbird}}: {{A Tailless Flapping Wing Micro Air Vehicle}}},
  shorttitle = {Development of the {{Nano Hummingbird}}},
  booktitle = {50th {{AIAA Aerospace Sciences Meeting}} Including the {{New Horizons Forum}} and {{Aerospace Exposition}}},
  author = {Keennon, Matthew and Klingebiel, Karl and Won, Henry},
  year = 2012,
  month = jan,
  publisher = {{American Institute of Aeronautics and Astronautics}},
  address = {Nashville, Tennessee},
  doi = {10.2514/6.2012-588},
  urldate = {2025-01-09},
  isbn = {978-1-60086-936-5},
  langid = {english}
}

@article{Nguyen2017,
  title = {Experimental {{Investigation}} of {{Wing Flexibility}} on {{Force Generation}} of a {{Hovering Flapping Wing Micro Air Vehicle}} with {{Double Wing Clap-and-Fling Effects}}},
  author = {Nguyen, Quoc V and Chan, Woei L and Debiasi, Marco},
  year = 2017,
  month = sep,
  journal = {International Journal of Micro Air Vehicles},
  volume = {9},
  number = {3},
  pages = {187--197},
  issn = {1756-8293, 1756-8307},
  doi = {10.1177/1756829317695565},
  urldate = {2026-04-10},
  langid = {english}
}

@article{Nguyen2018,
  title = {Development and {{Flight Performance}} of a {{Biologically-Inspired Tailless Flapping-Wing Micro Air Vehicle}} with {{Wing Stroke Plane Modulation}}},
  author = {Nguyen, Quoc-Viet and Chan, Woei Leong},
  year = 2018,
  month = dec,
  journal = {Bioinspiration \& Biomimetics},
  volume = {14},
  number = {1},
  pages = {016015},
  issn = {1748-3190},
  doi = {10.1088/1748-3190/aaefa0},
  urldate = {2025-02-12},
  langid = {english}
}

@article{Phan2017,
  title = {Design and {{Stable Flight}} of a 21 g {{Insect-Like Tailless Flapping Wing Micro Air Vehicle}} with {{Angular Rates Feedback Control}}},
  author = {Phan, Hoang Vu and Kang, Taesam and Park, Hoon Cheol},
  year = 2017,
  month = apr,
  journal = {Bioinspiration \& Biomimetics},
  volume = {12},
  number = {3},
  pages = {036006},
  issn = {1748-3190},
  doi = {10.1088/1748-3190/aa65db},
  urldate = {2025-02-03},
  langid = {english}
}

@article{Phan2017a,
  title = {An {{Experimental Comparative Study}} of the {{Efficiency}} of {{Twisted}} and {{Flat Flapping Wings During Hovering Flight}}},
  author = {Phan, Hoang Vu and Truong, Quang Tri and Park, Hoon Cheol},
  year = 2017,
  month = apr,
  journal = {Bioinspiration \& Biomimetics},
  volume = {12},
  number = {3},
  pages = {036009},
  issn = {1748-3190},
  doi = {10.1088/1748-3190/aa65e6},
  urldate = {2026-06-04},
  langid = {english}
}

@article{Phan2019,
  title = {Insect-{{Inspired}}, {{Tailless}}, {{Hover-Capable Flapping-Wing Robots}}: {{Recent Progress}}, {{Challenges}}, and {{Future Directions}}},
  shorttitle = {Insect-{{Inspired}}, {{Tailless}}, {{Hover-Capable Flapping-Wing Robots}}},
  author = {Phan, Hoang Vu and Park, Hoon Cheol},
  year = 2019,
  month = nov,
  journal = {Progress in Aerospace Sciences},
  volume = {111},
  pages = {100573},
  issn = {03760421},
  doi = {10.1016/j.paerosci.2019.100573},
  urldate = {2025-02-10},
  langid = {english}
}

@article{Pornsin-Sirirak2001,
  title = {Microbat: {{A Palm-Sized Electrically Powered Ornithopter}}},
  author = {{Pornsin-Sirirak}, T Nick and Tai, Yu-Chong and Ho, Chih-Ming and Keennon, Matt},
  year = 2001,
  journal = {Proceedings of NASA/JPL workshop on biomorphic robotics},
  volume = {14},
  pages = {17},
  langid = {english}
}

@article{Preumont2021,
  title = {A {{Note}} on the {{Electromechanical Design}} of a {{Robotic Hummingbird}}},
  author = {Preumont, Andr{\'e} and Wang, Han and Kang, Shengzheng and Wang, Kainan and Roshanbin, Ali},
  year = 2021,
  month = mar,
  journal = {Actuators},
  volume = {10},
  number = {3},
  pages = {52},
  issn = {2076-0825},
  doi = {10.3390/act10030052},
  urldate = {2025-02-05},
  copyright = {https://creativecommons.org/licenses/by/4.0/},
  langid = {english}
}

@article{RafeeNekoo2025,
  title = {A {{Review}} on {{Flapping-Wing Robots}}: {{Recent Progress}} and {{Challenges}}},
  shorttitle = {A {{Review}} on {{Flapping-Wing Robots}}},
  author = {Rafee Nekoo, Saeed and Rashad, Ramy and De Wagter, Christophe and Fuller, Sawyer B and Croon, Guido De and Stramigioli, Stefano and Ollero, Anibal},
  year = 2025,
  month = may,
  journal = {The International Journal of Robotics Research},
  pages = {02783649251343638},
  issn = {0278-3649, 1741-3176},
  doi = {10.1177/02783649251343638},
  urldate = {2025-08-08},
  langid = {english}
}

@article{Roshanbin2017,
  title = {Colibri: {{A Hovering Flapping Twin-Wing Robot}}},
  shorttitle = {Colibri},
  author = {Roshanbin, A and Altartouri, H and Kar{\'a}sek, M and Preumont, A},
  year = 2017,
  month = dec,
  journal = {International Journal of Micro Air Vehicles},
  volume = {9},
  number = {4},
  pages = {270--282},
  issn = {1756-8293, 1756-8307},
  doi = {10.1177/1756829317695563},
  urldate = {2025-02-05},
  langid = {english}
}

@article{Sane2002,
  title = {The {{Aerodynamic Effects}} of {{Wing Rotation}} and a {{Revised Quasi-Steady Model}} of {{Flapping Flight}}},
  author = {Sane, Sanjay P. and Dickinson, Michael H.},
  year = 2002,
  month = apr,
  journal = {Journal of Experimental Biology},
  volume = {205},
  number = {8},
  pages = {1087--1096},
  issn = {1477-9145, 0022-0949},
  doi = {10.1242/jeb.205.8.1087},
  urldate = {2026-04-20},
  langid = {english}
}

@article{Sane2003,
  title = {The {{Aerodynamics}} of {{Insect Flight}}},
  author = {Sane, Sanjay P.},
  year = 2003,
  month = dec,
  journal = {Journal of Experimental Biology},
  volume = {206},
  number = {23},
  pages = {4191--4208},
  issn = {1477-9145, 0022-0949},
  doi = {10.1242/jeb.00663},
  urldate = {2025-08-07},
  langid = {english}
}

@article{Thomas2001,
  title = {Animal {{Flight Dynamics I}}. {{Stability}} in {{Gliding Flight}}},
  author = {Thomas, Adrian L.R. and Taylor, Graham K.},
  year = 2001,
  month = oct,
  journal = {Journal of Theoretical Biology},
  volume = {212},
  number = {3},
  pages = {399--424},
  issn = {00225193},
  doi = {10.1006/jtbi.2001.2387},
  urldate = {2026-04-21},
  langid = {english}
}

@article{Tu2020,
  title = {Untethered {{Flight}} of an At-{{Scale Dual-Motor Hummingbird Robot}} with {{Bio-Inspired Decoupled Wings}}},
  author = {Tu, Zhan and Fei, Fan and Deng, Xinyan},
  year = 2020,
  journal = {IEEE Robotics and Automation Letters},
  pages = {1--1},
  issn = {2377-3766, 2377-3774},
  doi = {10.1109/LRA.2020.2974717},
  urldate = {2025-02-10},
  copyright = {https://ieeexplore.ieee.org/Xplorehelp/downloads/license-information/IEEE.html},
  langid = {english}
}

@article{Weis-Fogh1973,
  title = {Quick {{Estimates}} of {{Flight Fitness}} in {{Hovering Animals}}, {{Including Novel Mechanisms}} for {{Lift Production}}},
  author = {{Weis-Fogh}, Torkel},
  year = 1973,
  month = aug,
  journal = {Journal of Experimental Biology},
  volume = {59},
  number = {1},
  pages = {169--230},
  issn = {0022-0949, 1477-9145},
  doi = {10.1242/jeb.59.1.169},
  urldate = {2026-04-21},
  copyright = {http://www.biologists.com/user-licence-1-1/},
  langid = {english}
}

@article{Wu2024,
  title = {A {{Multi-Modal Tailless Flapping-Wing Robot Capable}} of {{Flying}}, {{Crawling}}, {{Self-Righting}} and {{Horizontal Take-Off}}},
  author = {Wu, Chaofeng and Xiao, Yiming and Zhao, Jiaxin and Mou, Jiawang and Cui, Feng and Liu, Wu},
  year = 2024,
  month = may,
  journal = {IEEE Robotics and Automation Letters},
  volume = {9},
  number = {5},
  pages = {4734--4741},
  issn = {2377-3766, 2377-3774},
  doi = {10.1109/LRA.2024.3384910},
  urldate = {2025-02-19},
  copyright = {https://ieeexplore.ieee.org/Xplorehelp/downloads/license-information/IEEE.html},
  langid = {english}
}

@article{Wu2026,
  title = {Design and {{Experimental Validation}} of a {{Tailless Flapping-Wing Micro Aerial Vehicle}} with {{Long Endurance}} and {{High Payload Capability}}},
  author = {Wu, Chaofeng and Xiao, Yiming and Zhao, Jiaxin and Guo, Qingcheng and Cui, Feng and Wu, Xiaosheng and Liu, Wu},
  year = 2026,
  month = jan,
  journal = {Drones},
  volume = {10},
  number = {1},
  pages = {26},
  issn = {2504-446X},
  doi = {10.3390/drones10010026},
  urldate = {2026-02-24},
  langid = {english}
}

@article{Zufferey2022,
  title = {How {{Ornithopters Can Perch Autonomously}} on a {{Branch}}},
  author = {Zufferey, Raphael and {Tormo-Barbero}, Jesus and {Feliu-Taleg{\'o}n}, Daniel and Nekoo, Saeed Rafee and Acosta, Jos{\'e} {\'A}ngel and Ollero, Anibal},
  year = 2022,
  month = dec,
  journal = {Nature Communications},
  volume = {13},
  number = {1},
  pages = {7713},
  issn = {2041-1723},
  doi = {10.1038/s41467-022-35356-5},
  urldate = {2026-04-10},
  langid = {english}
}

\end{document}